\documentclass{ecai} 
\usepackage{enumitem}
\usepackage{graphicx}
\usepackage{color}
\usepackage{amsmath}
\usepackage{amsfonts}
\usepackage{tcolorbox}
\usepackage{framed}
\usepackage{url}

\renewenvironment{leftbar}[1][\hsize]
{%
  \MakeFramed{\hsize#1\advance\hsize-\width\FrameRestore\vskip-1pt}
}
{\vskip-3pt\endMakeFramed} 
\makeatletter
\renewenvironment{figure}
  {\@float{figure}}
  {\end@float}
\makeatother

\def \eg{\emph{e.g.}}
\def \ie{\emph{i.e.}}

\def \kanon{$k$-anonymity}
\def \ldiv{$\ell$-diversity}
\def \tclos{$t$-closeness}
\def \acsincome{\small{\textsf{ACSIncome}}}
\def \compas{\small{\textsf{Compas}}}
\def \adult{\small{\textsf{Adult}}}
\def \race{\small{\texttt{race}}}
\def \gender{\small{\texttt{gender}}}

\newcommand{\BibTeX}{B\kern-.05em{\sc i\kern-.025em b}\kern-.08em\TeX}

\begin{document}
\sloppy

\begin{frontmatter}


\paperid{3224} 


\title{Fair Play for Individuals, Foul Play for Groups?\\Auditing Anonymization's Impact on ML Fairness}


\author[A]{\fnms{Héber H.}~\snm{Arcolezi}}
\author[B]{\fnms{Mina}~\snm{Alishahi}\thanks{Corresponding Author. Email: mina.sheikhalishahi@ou.nl}}
\author[C]{\fnms{Adda-Akram}~\snm{Bendoukha}} 
\author[C]{\fnms{Nesrine}~\snm{Kaaniche}} 

\address[A]{Inria Centre at the University Grenoble Alpes, France}
\address[B]{Open Universiteit, Netherlands}
\address[C]{Samovar, Télécom SudParis, Institut Polytechnique de Paris, France}


\begin{abstract}
Machine learning (ML) algorithms are heavily based on the availability of training data, which, depending on the domain, often includes sensitive information about data providers. This raises critical privacy concerns. Anonymization techniques have emerged as a practical solution to address these issues by generalizing features or suppressing data to make it more difficult to accurately identify individuals. Although recent studies have shown that privacy-enhancing technologies can influence ML predictions across different subgroups, thus affecting fair decision-making, the specific effects of anonymization techniques, such as $k$-anonymity, $\ell$-diversity, and $t$-closeness, on ML fairness remain largely unexplored. In this work, we systematically audit the impact of anonymization techniques on ML fairness, evaluating both individual and group fairness. Our quantitative study reveals that anonymization can degrade group fairness metrics by up to fourfold. Conversely, similarity-based individual fairness metrics tend to improve under stronger anonymization, largely as a result of increased input homogeneity. By analyzing varying levels of anonymization across diverse privacy settings and data distributions, this study provides critical insights into the trade-offs between privacy, fairness, and utility, offering actionable guidelines for responsible AI development. Our code is publicly available at: \url{https://github.com/hharcolezi/anonymity-impact-fairness}.
\end{abstract}

\end{frontmatter}

\section{Introduction}

As machine learning (ML) systems increasingly shape critical decision-making across domains such as healthcare, finance, and social services, concerns about \textit{privacy} and \textit{fairness} have gained significant attention. Privacy is essential to safeguard personal data, ensuring compliance with regulatory frameworks and protecting individuals from potential data misuse. Fairness, on the other hand, ensures that ML models provide unbiased and equitable outcomes across different demographic groups. Both aspects are fundamental to fostering trust and accountability in AI-driven decision-making.

The growing emphasis on privacy is reflected in strict regulatory frameworks such as the {General Data Protection Regulation} (GDPR)~\cite{GDPR2016} and the {California Consumer Privacy Act} (CCPA)~\cite{CCPA2018}, which impose stringent requirements on data collection, storage, and sharing. Beyond privacy, the proposed European Union AI Act~\cite{AIACT2024} extends regulatory concerns to fairness, requiring AI systems to be transparent, non-discriminatory, and aligned with ethical and social values. These evolving legal frameworks highlight the dual imperative of privacy and fairness in the design of  ML models.

Given the importance of balancing privacy and fairness in machine learning, a substantial body of work has investigated their interplay—primarily through the lens of differential privacy (DP)~\cite{yao2025sok,Fioretto_2022}. Much of this literature focuses on central DP mechanisms~\cite{Dwork2006}, such as DP-SGD, which have been shown to exacerbate group fairness disparities in some settings~\cite{Bagdasaryan2019}, while other studies report more bounded effects~\cite{deoliveira2024}. On the other hand, few works have concluded that Local DP mechanisms positively impact group fairness metrics~\cite{Arcolezi_2023,Carey2023,makhlouf2024impact,makhlouf2024systematic} by removing the dependency between protected attributes and the target variable. 

Despite growing research on the interplay between fairness and DP, the \emph{fairness implications of anonymized datasets on ML remain unexplored}. Anonymization methods, such as $k$-anonymity~\cite{Samarati2001,SWEENEY2002}, provide privacy guarantees by generalizing specific attributes or suppressing data to prevent re-identification attacks~\cite{Machanavajjhala2007, Li2007, 8890382}. These techniques are widely used due to their simplicity, interpretability~\cite{GaballahAANZMM24, FATHALIZADEH2022102665}, and compatibility with existing privacy regulations such as GDPR, as noted by the Article 29 Working Party~\cite{WP29_2014}.

Ensuring fairness in ML involves mainly two key principles: (i) \textit{group fairness}, which ensures that model predictions are consistent across demographic groups~\cite{10.1145/3616865, Barocas2023}, and (ii) \textit{individual fairness}, which ensures that similar individuals receive similar treatment~\cite{Dwork2012}. While anonymization techniques are effective in preserving privacy, they introduce transformations such as generalization or suppression, which may distort data distributions and induce unintended bias in ML models~\cite{10.1007/978-3-030}. These alterations can inadvertently affect protected attributes, shift group distributions, and influence fairness metrics.
\textcolor{black}{We hypothesize that such statistical alterations introduced to satisfy anonymization can significantly impact fairness metrics in ML.}

Existing research on the interplay between anonymization and fairness has primarily focused on \textit{dataset-level fairness}, evaluating how anonymization techniques alter dataset properties~\cite{2870623}. Some studies have explored optimal parameter selection, such as determining the best value of $t$ in \textit{$t$-closeness}, to balance privacy and fairness trade-offs~\cite{Hajian2012, 6754013}. However, \textit{a significant gap remains in understanding the direct impact of anonymization on model fairness}, particularly how these techniques influence bias propagation and fairness metrics in ML models. Addressing this challenge is crucial for the development of privacy-preserving yet fair AI systems.

\textbf{Our contributions.} In this paper, we present the first in-depth, systematic audit of the impact of three widely-used anonymization methods, namely, \kanon~\cite{Samarati2001,SWEENEY2002}, \ldiv~\cite{Machanavajjhala2007}, and \tclos{}~\cite{Li2007} on fairness in ML. We analyze their effects on both \textit{group fairness metrics} (\eg, equal opportunity~\cite{hardt2016equality}), \textit{individual fairness metrics} (\eg, similarity fairness~\cite{Dwork2012}), while also examining the trade-offs with \textit{utility metrics} (\eg, F1-score). Our findings highlight the nuanced interplay between anonymity, fairness, and utility, offering valuable insights and actionable guidance for practitioners working with anonymized datasets.

In summary, the key contributions of this paper are: 

\begin{itemize} 
    \item We conduct a comprehensive audit on the effects of \kanon, \ldiv, and \tclos{} on both group and individual fairness metrics across diverse datasets and ML models. 
    {This dual focus offers a nuanced understanding of how anonymization techniques influence different aspects of fairness.}
    
    \item {We analyze how factors such as record suppression thresholds, dataset size, and target class balance modulate the fairness-utility trade-off. This enables a fine-grained understanding of how anonymization interacts with data characteristics in practice.}
    
    \item Based on our findings, we provide practical guidelines for mitigating fairness risks in ML pipelines. These include strategies for balancing privacy and fairness under different anonymization configurations, empowering practitioners to make informed decisions when deploying privacy-preserving technologies.
\end{itemize}

\textbf{Outline.} The remainder of this paper is structured as follows.
In Section~\ref{sec:preliminaries}, we present preliminary concepts. 
In Section~\ref{sec:methodology}, we formulate the problem, describe our research questions, and present our experimental setup.
Afterward, in Section~\ref{sec:results}, we present the experimental results with an analysis and guidelines for practitioners, and we conclude the paper with a discussion for future work in Section~\ref{sec:conclusion}.

\section{Preliminaries} \label{sec:preliminaries}

\subsection{Anonymization Methods} \label{sub:anonymization_methods}

Three main anonymization methods are commonly used: \kanon~\cite{Samarati2001,SWEENEY2002}, \ldiv~\cite{Machanavajjhala2007}, and \tclos~\cite{Li2007}. These methods aim to reduce the risk of re-identification by transforming the data while maintaining its utility for downstream tasks. 
In the context of anonymization, attributes in a dataset are typically categorized as follows:
1) \textit{identifiers} that uniquely identify individuals, such as social security numbers, 
2) \textit{quasi-identifier} (QI) attributes that, when combined, may potentially identify an individual, such as zip codes, birth dates, and genders. { While these attributes lack uniqueness in isolation, their conjunction often yields a one-to-one mapping with identifiers.} And 
3) \textit{sensitive} attributes that adversaries are prohibited from discovering, such as a patient's disease or an employer's salary. An equivalence class, also known as an equivalence group, is a set of records in a dataset that share the same values for the quasi-identifier attributes. 
We now formally define each anonymization method.

\textbf{$k$-anonymity:} A dataset \( D \) satisfies \( k \)-anonymity with respect to a set of $QIs$, if and only if every equivalence class \( E \subseteq D \) formed by the unique combinations of values in \( QI \) contains at least \( k \) records~\cite{Samarati2001,SWEENEY2002}. Formally,
$ \forall E \subseteq D, \ |E| \geq k$,  where \( E \) is an equivalence class of records that share the same generalized values across \( QI \). The \kanon{} property ensures that any individual represented in the dataset cannot be distinguished from at least \( k-1 \) other individuals based on the quasi-identifier attributes, thus reducing the risk of re-identification. Achieving \kanon{} typically involves generalization (replacing specific values with broader categories) or suppression of quasi-identifiers to create equivalence classes.

\textbf{$\ell$-diversity:} A dataset \( D \) satisfies \( \ell \)-diversity with respect to a set of attributes \( QIs \) and a sensitive attribute \( S \) if, for every equivalence class \( E \subseteq D \) formed by unique combinations of values in \( QI \), there are at least \( \ell \) well-represented distinct values of \( S \)~\cite{Machanavajjhala2007}. Formally, $ \forall E \subseteq D, \ | \text{distinct}(S(E)) | \geq \ell$, where \( S(E) \) denotes the set of sensitive attribute values in the equivalence class \( E \). \ldiv{} ensures that each equivalence class contains sufficient diversity in the sensitive attribute values, mitigating the risk of attribute disclosure. 

\textbf{$t$-closeness:}
A dataset \( D \) satisfies \( t \)-closeness with respect to a set of attributes \( QI \) and a sensitive attribute \( S \) if, for every equivalence class \( E \subseteq D \), the distribution of \( S \) within \( E \) is within a distance \( t \) from the distribution of \( S \) in the overall dataset \( D \)~\cite{Li2007}. Formally, 
$\forall E \subseteq D, \ d(S(E), S(D)) \leq t$,
 where \( d \) is a distance metric, such as the Earth Mover’s Distance (EMD)~\cite{rubner2000earth}, and \( S(E) \) and \( S(D) \) represent the distributions of \( S \) in \( E \) and \( D \), respectively.
The \tclos{} property ensures that sensitive attribute distributions in each equivalence class closely resemble the overall distribution, reducing the risk of inference attacks. Generalization and suppression are applied to satisfy the \( t \)-closeness constraint.

\subsection{ML Fairness Metrics} \label{sub:fairness_metrics}
Machine learning fairness~\cite{Barocas2023} refers to the principle that ML models should produce predictions or decisions that are impartial and equitable across individuals or groups, particularly when protected attributes such as race, gender, or socioeconomic status are involved. This study employs a comprehensive evaluation of fairness using two primary categories of metrics: group fairness metrics (Section~\ref{sub:group_fairness}) and individual fairness metrics (Section~\ref{sub:individual_fairness}). \textcolor{black}{Their joint use allows us to examine the multifaceted impact of anonymization on both group-level disparities and individual-level consistency.}

\subsubsection{Group fairness metrics.} \label{sub:group_fairness}
Group fairness metrics assess the degree to which model outcomes are distributed equitably across demographic groups defined by protected attributes. These metrics aim to ensure that the treatment of groups is consistent and does not result in discrimination. 

\textbf{Model Accuracy Difference (MAD)} quantifies the difference in model accuracy between two specific demographic groups defined by a protected attribute $A$. Specifically, MAD evaluates whether the model performs equally well for individuals in group \( A = 1 \) compared to those in group \( A = 0 \). Formally:
\begin{equation} \label{eq:mad}
    \text{MAD} = \Pr\left[\hat{Y} = Y \mid A = 1\right] - \Pr\left[\hat{Y} = Y \mid A = 0\right] 
\end{equation}
where \( Y \) is the true label, \( \hat{Y} \) is the predicted label, and \( A \) is the binary protected attribute defining two groups (\( A = 1 \) and \( A = 0 \)).
A value of \( \text{MAD} = 0 \) signifies equal model accuracy across the two groups.

\textbf{\textcolor{black}{Equal Opportunity Difference} (EOD)}~\cite{hardt2016equality} requires that predictions are independent of the protected attribute \emph{conditional on the positive class only}
(\ie, $\hat{Y} \perp A \mid Y{=}1$). It measures the disparity in \emph{true positive rates} across groups.
Formally:
\begin{equation} \label{eq:eod}
    \text{EOD} = \Pr\left[\hat{Y} = 1 \mid Y = 1, A = 1\right] - \Pr\left[\hat{Y} = 1 \mid Y = 1, A = 0\right]
\end{equation}
An \( \text{EOD} = 0 \) indicates equality of true positive rates across groups.

\textbf{Statistical Parity Difference (SPD)}~\cite{Dwork2012} evaluates whether the likelihood of a positive outcome is the same across two specific demographic groups defined by a protected attribute. Formally:
\begin{equation} \label{eq:spd}
    \text{SPD} = \Pr \left[ \hat{Y} = 1 \mid A = 1 \right] - \Pr\left[\hat{Y} = 1 \mid A = 0\right]
\end{equation}
An \( \text{SPD} = 0 \) indicates equal selection rates between the groups.

\subsubsection{Individual fairness metrics.} \label{sub:individual_fairness}
Individual fairness metrics evaluate the consistency of a model’s predictions for similar individuals~\cite{Dwork2012}. 

\textbf{Lipschitz Fairness (LF)} evaluates the maximum sensitivity of a model's predictions to changes in its input features. It measures the largest rate of change in the model’s outputs relative to input variations, quantified by the Lipschitz constant. A lower LF value indicates better fairness, as it reflects reduced sensitivity and ensures that similar inputs yield similar predictions.
Formally:
\begin{equation} \label{eq:lf}
    \text{LF} =     \max_{i \neq j} \frac{\text{diff}(f(x_i), f(x_j))}{\text{dist}(x_i, x_j)} \mathrm{,}
\end{equation}
\noindent where \( f(x) \) is the model's prediction for input \( x \), \( \text{dist}(x_i, x_j) \) is the distance between inputs \( x_i \) and \( x_j \), and \( \text{diff}(f(x_i), f(x_j)) \) is the difference between the model's predictions for inputs \( x_i \) and \( x_j \), often computed using measures such as entropy for classification tasks.

\textbf{Similarity Fairness (SF)} quantifies the \emph{local smoothness} of the model: within each point’s neighborhood, it aggregates absolute prediction differences and weights them by the feature-space distance between pairs. This distance-weighted variant emphasizes consistency across the broader local region while de-emphasizing near-duplicate pairs. Lower SF values indicate smoother, more locally consistent predictions.
Formally:
\begin{equation} \label{eq:sf}
    \text{SF} = \frac{1}{n} \sum_{i=1}^{n} \frac{1}{|\mathcal{N}(x_i)|} \sum_{x_j \in \mathcal{N}(x_i)} \left| f(x_i) - f(x_j) \right| \cdot \text{dist}(x_i, x_j) \mathrm{,}
\end{equation}
\noindent where \( \mathcal{N}(x_i) \) represents the neighborhood of \( x_i \), \( \text{dist}(x_i, x_j) \) is the distance between $x_i$ and $x_j$ in feature space, and \( n \) is the total number of data points.

\textbf{Neighborhood Consistency Fairness (NCF)} evaluates the consistency of predictions within local neighborhoods, ensuring equitable treatment for instances with similar characteristics~\cite{Zemel2013}. 
A lower NCF value reflects better fairness, as it ensures that predictions are consistent for neighboring inputs.
Formally:
\begin{equation} \label{eq:ncf}
    \text{NCF} = \frac{1}{n} \sum_{i=1}^{n} \frac{1}{|\mathcal{N}(x_i)|} \sum_{x_j \in \mathcal{N}(x_i)} \mathbb{I}\left( f(x_i) \neq f(x_j) \right) \mathrm{,}
\end{equation}
\noindent where \( \mathcal{N}(x_i) \) is the neighborhood of \( x_i \),  \( \mathbb{I}(\cdot) \) is the indicator function, which equals 1 if predictions for \( x_i \) and \( x_j \) are mismatched, and \( n \) is the total number of data points.

{
\textbf{Approximation via \( k \)-Nearest Neighbors (k-NN):} Computing LF, SF, and NCF requires evaluating pairwise distances between all data points, resulting in a computational complexity of \( O(n^2) \), which is prohibitive for large datasets. To improve efficiency, we approximate these metrics using k-NN, reducing the complexity to \( O(kn) \), where \( k \ll n \), as also motivated in prior work~\cite{Zemel2013}. Specifically, each sample is compared with its $k$-nearest neighbors instead of the entire dataset.} We use Euclidean distance and set \( k = 100 \), which provides a balance between computational tractability and the ability to capture a meaningful local neighborhood. Throughout the paper, we refer to these approximated versions as Approximate LF (ALF) and Approximate SF (ASF).
 
\section{Methodology and Experimental Setup} \label{sec:methodology}

This section formulates the problem, introduces the research questions, and describes the experimental setup.

\subsection{Problem Formulation} \label{sub:problem_formulation}

We aim to systematically audit how different anonymization methods influence group and individual fairness outcomes, as well as predictive performance, across a range of ML models and datasets.

To formalize this problem, we define the following components. 
Let \( \mathcal{D} = \{(a_i, x_i, y_i)\}_{i=1}^n \) be a dataset consisting of \( n \) i.i.d. samples drawn from an unknown joint distribution over \( \mathcal{A} \times \mathcal{X} \times \mathcal{Y} \), where:

\begin{itemize}[leftmargin=*]
    \item \( A \in \mathcal{A} \): Protected attribute(s) (\eg, race, gender), \textcolor{black}{treated as quasi-identifiers for anonymization and explicitly included as input features. We adopt this approach in line with prior literature showing that excluding protected attributes (\ie, ``fairness through unawareness'') does not prevent discriminatory outcomes~\cite{Barocas2023,kusner2017counterfactual,10.1145/3616865}. We focus on binary protected attributes to ensure comparability across standard fairness metrics and to align with common preprocessing practices in the fairness and privacy literature~\cite{yao2025sok}.}
    
    \item \( X \in \mathcal{X} \): Non-sensitive, non-protected features (\eg, employment status, credit score)\textcolor{black}{, which are also treated as quasi-identifiers, thus, subject to generalization or suppression.}
    
    \item \( Y \in \{0,1\} \): Binary target variable (\eg, loan approval outcome), where \( Y = 1 \) denotes a favorable decision. \textcolor{black}{We focus on binary targets for interpretability and comparability across fairness metrics, consistent with standard practices in the fair-ML~\cite{Zemel2013,Barocas2023} and privacy-fairness literature~\cite{Bagdasaryan2019,Arcolezi_2023,Carey2023,deoliveira2024,makhlouf2024systematic}.}
\end{itemize}

Given this setup, a supervised ML model learns a predictive function \( f: \mathcal{A} \times \mathcal{X} \rightarrow [0,1] \) such that \( \hat{Y} = f(A, X) \) approximates the true label \( Y \). 
In practice, the model is trained on a dataset \( D' \), which may differ from \( D \) due to privacy constraints.
Specifically, to preserve privacy, we consider an anonymization mechanism \( \mathcal{M} \), which transforms the raw dataset \( D \) into an anonymized dataset \( D' = \mathcal{M}(D) \). 
In this work, \( \mathcal{M} \) represents one of the following methods: \kanon, \ldiv, or \tclos{}, each of which enforces privacy by applying generalization and suppression over quasi-identifiers. 
We then train ML classifiers on both original (\( D \)) and anonymized (\( D' \)) datasets, and compare them using performance (\eg, F1-score) and fairness metrics (Section~\ref{sub:fairness_metrics}).

\subsection{Research Questions} \label{sub:research_questions}
We structure our evaluation around the following research questions:

\begin{itemize}
    \item \textbf{RQ1)} \textit{How do different anonymization techniques and anonymity levels affect the fairness of ML models?} This research question explores the impact of three widely used anonymization techniques (\kanon, \ldiv, and \tclos) on the fairness of ML models. By adjusting their respective privacy parameters (\ie, $k$, $\ell$, and $t$), we evaluate how these techniques influence fairness metrics and whether certain configurations disproportionately affect specific demographic groups.  
    The experiments addressing this question are presented in Section~\ref{sub:main_results}.
    
    \item \textbf{RQ2)} \textit{What is the impact of varying the record level suppression in anonymization on the fairness of ML models?} Because suppression often targets outlier data, this may disproportionately affect certain sub-populations, potentially exacerbating fairness disparities. This research question investigates how varying suppression thresholds (removing rows) impact fairness metrics. The experiments addressing this question are detailed in Section~\ref{sub:results_suppression}.

    \item \textbf{RQ3)} \textit{What is the impact of varying target distributions on the fairness of ML models?} This research question examines how changes in the target distribution, specifically by varying the threshold used to binarize the \textit{target variable} (see "Datasets" in Section~\ref{sub:setup_experiments}), influence fairness metrics. \textcolor{black}{In many real-world settings, continuous outcomes (\eg, income, risk score) are binarized to support binary decision-making processes such as loan approvals, benefit eligibility, or pretrial release decisions. However, the choice of threshold can significantly affect the proportion of positive versus negative labels, altering group-specific outcome rates and thereby impacting fairness metrics.} The experiments addressing this question are presented in Section~\ref{sub:results_target_distribution}.
    
    \item \textbf{RQ4)} \textit{How does dataset size influence the relationship between anonymization and fairness in ML models?} This question explores the role of dataset size in mediating the relationship between anonymization and fairness. 
    \textcolor{black}{In practice, smaller datasets may be more vulnerable to the distortions introduced by anonymization, such as over-generalization or excessive suppression, which could disproportionately affect minority groups or rare feature combinations.}
    By systematically varying the data fraction, we analyze how sample size influences the trade-offs between privacy, fairness, and utility. The experiments addressing this question are detailed in Section~\ref{sub:results_data_size}.
    
    \item \textbf{RQ5)} \textit{To what extent are the fairness results obtained with XGBoost representative across different ML classifiers?} This research question investigates whether the fairness results observed in our default experiments using XGBoost~\cite{XGBoost} (\textbf{RQ1} -- \textbf{RQ4}) generalize across other ML classifiers. By comparing the fairness metrics and predictive performance of multiple classifiers (\eg, Random Forest, Neural Networks), we aim to assess whether the trends observed with XGBoost are consistent. The experiments addressing this question are presented in Section~\ref{sub:results_ml_classifier}.
\end{itemize}

\subsection{Experimental Setup} \label{sub:setup_experiments}

\textbf{Environment:} All algorithms are implemented in Python3 and executed on a local machine with 2.50GHz Intel Core i9 and 64GB RAM. The source code is publicly available in our \textit{\textbf{GitHub repository:}} \url{https://github.com/hharcolezi/anonymity-impact-fairness}.

\textbf{Datasets:}
{For our experiments, we used three widely-used benchmark datasets in the fairness literature: the \adult{} dataset from the UCI ML repository~\cite{ADULT}, the \compas{} dataset gathered by ProPublica~\cite{compas}, and the \acsincome{} retrieved with the \texttt{folktables}~\cite{ding2021retiring} Python library. The datasets are randomly split into a training set ($80\%$) and a testing set ($20\%$). To simulate a worst-case privacy scenario, all attributes ($X$ and $A$) are treated as quasi-identifiers subject to generalization or suppression during anonymization. For fairness evaluation, we consider both \gender{} and \race{} as protected attributes across all datasets. A detailed description of each dataset is provided in Appendix~\ref{app:datasets}.}

\textbf{Anonymization parameters:} The anonymization parameters are varied as follows: $k \in \{1, 2, \ldots, 9, 10, 25, 50, 75, 100\}$ for \kanon, $\ell=2$ for \ldiv{} (binary target), and $t \in \{0.45, 0.50, 0.55\}$ for \tclos\footnote{{The values of 
$t$ specify upper bounds on the Earth Mover's Distance between the distribution of the sensitive attribute in the entire dataset and its distribution within each equivalence class.}}. Note that for $k=1$, no anonymity is satisfied, which serves as the \textit{non-private baseline}. Unless otherwise mentioned, the allowed record suppression level is fixed at $\textsf{supp\_level}=20\%$\footnote{{Our attempts with lower suppression levels (0\%-15\%) did not achieve anonymization for high $k$ values because we operated under a \emph{worst-case} assumption where all attributes except the target were treated as QIDs.}}. The anonymization methods were implemented with Anjana~\cite{sainz2024open}. The same generalization levels applied to the training set are replicated for the test set to ensure consistency and prevent discrepancies between training and evaluation.

\textbf{Model training:} For our experiments, we use XGBoost~\cite{XGBoost} as the default ML classifier. In Section~\ref{sub:results_ml_classifier}, we further benchmark the performance and fairness of other state-of-the-art ML classifiers, including LightGBM (LGBM)~\cite{LGBM}, Random Forests~\cite{Breiman2001}, and Neural Networks (\ie, Multi-Layer Perceptron -- MLP). Classifiers are trained using their default hyperparameters to ensure consistency across experiments. All models are trained and evaluated on both the original and anonymized datasets. Evaluation metrics are computed on predictions from the transformed test sets, enabling a comparative analysis of fairness and utility under different anonymization scenarios.

\textbf{Metrics:} We evaluate the performance of ML models trained on the original data (\ie, baseline $k=1$) and anonymized data on utility and fairness. First, for \textit{utility}, we use accuracy (ACC), F1-score (F1), and the area under the receiver operating characteristic curve (ROC AUC). Second, for \textit{fairness}, we assess group and individual fairness metrics as defined in Section~\ref{sub:fairness_metrics} (\ie, MAD, SPD, EOD, ALF, ASF, and NCF). To address the randomness in train-test splitting and ML algorithms, all experiments are repeated over 40 runs, with the results reported as averages alongside their standard deviations.

\section{Results and Analysis} \label{sec:results}

Following the methodology and experimental setup described in the previous Section~\ref{sec:methodology}, this section presents an analysis of the impact of anonymization methods on ML fairness. Due to space limitations, we present in the main paper the results obtained using the \adult{} dataset with \gender{} as the protected attribute. Additional details and discussions, including results with \race{} as the protected attribute and those for both \compas{} and \acsincome{} datasets (evaluating both \gender{} and \race{} as protected attributes), are provided in Appendix~\ref{app:add_results}. 
However, it is important to emphasize that the discussions in this section are broadly applicable to the findings across all datasets and all protected attributes.

\subsection{Impact of Anonymization on Fairness in ML} \label{sub:main_results}

\begin{figure*}[!htb]
    \centering
    \includegraphics[width=0.84\linewidth]{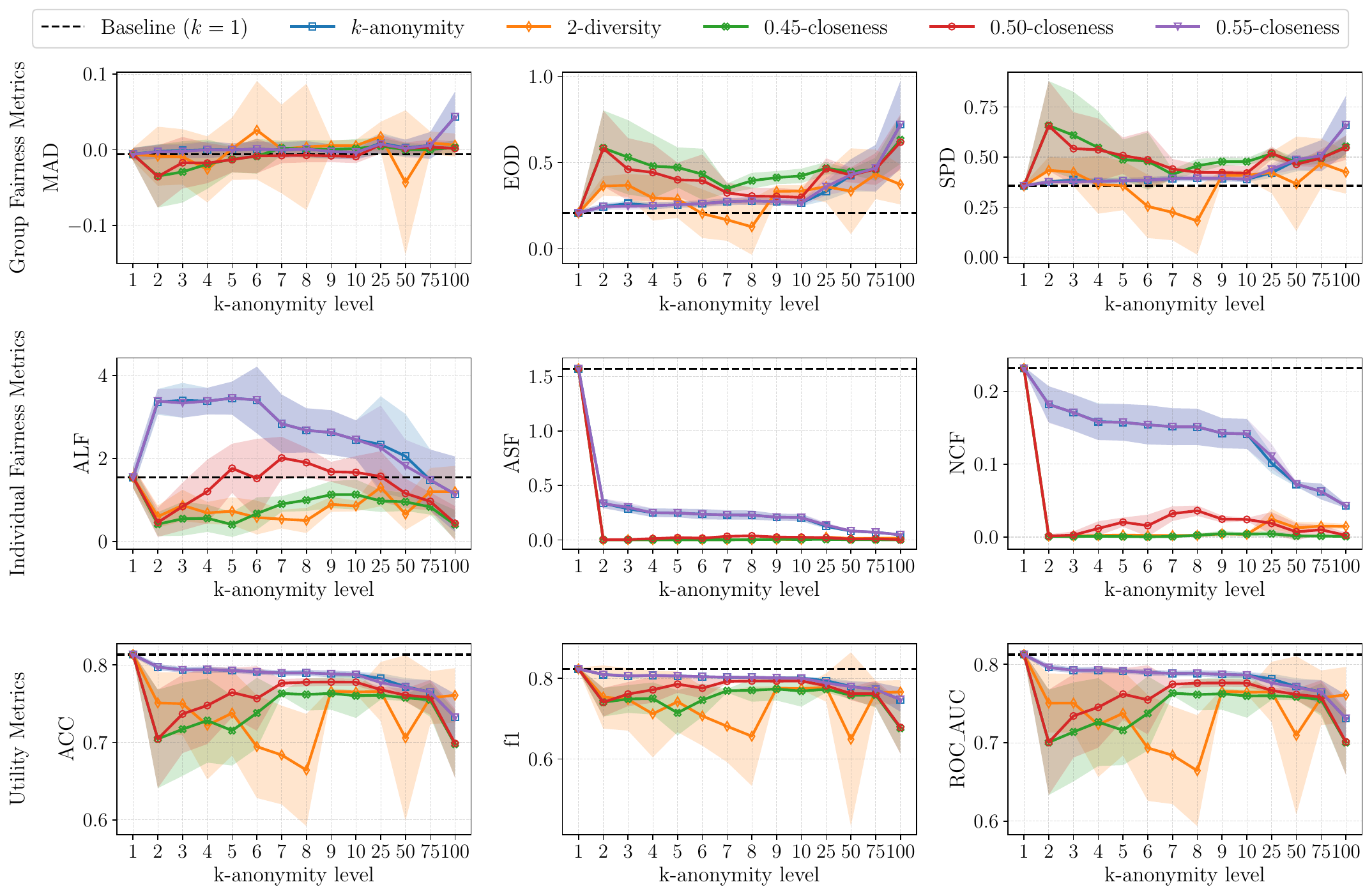}
    \caption{Impact of anonymity methods (\kanon, \ldiv, \tclos) on group fairness metrics (MAD, EOD, SPD), individual fairness metrics (ALF, ASF, NCF), and utility metrics (Accuracy, F1-score, ROC AUC) in ML. The results are based on the \adult{} dataset, with \gender{} as the protected attribute for fairness evaluation.}
    \label{fig:main_results_adult_gender}
\end{figure*}

To answer \textbf{RQ1} from Section~\ref{sub:research_questions}, we consistently analyze the impact of anonymity methods on fairness in ML. Specifically, Figure~\ref{fig:main_results_adult_gender} illustrates the impact of three anonymization techniques (\ie, \kanon, \ldiv, and \tclos) on group fairness (MAD, EOD, SPD), individual fairness (ALF, ASF, NCF), and utility (ACC, F1, ROC AUC) metrics, for the \adult{} dataset with \gender{} as the protected attribute. The results are aggregated across different privacy parameter levels to provide a comprehensive overview of the trade-offs among anonymity, fairness, and utility in ML models.

\paragraph{Group fairness.} From Figure~\ref{fig:main_results_adult_gender}, it is evident that \textit{anonymization techniques negatively affect group fairness metrics} such as SPD and EOD. For instance, as the parameter $k$ increases in \kanon, these fairness metrics tend to degrade, suggesting that stricter privacy constraints may amplify fairness disparities between demographic groups. 
Specifically, at $k=100$, the EOD metric increases from $\mathrm{EOD}=0.2$ (baseline $k=1$) to nearly $\mathrm{EOD}=0.8$ for both \kanon{} and \tclos, representing an approximately \textit{fourfold increase}. Similarly, the SPD metric rises from $\mathrm{SPD}=0.38$ (baseline $k=1$) to almost $\mathrm{SPD}=0.68$ (for \kanon{} and \tclos). 
\textcolor{black}{However, this degradation is \textit{not always monotonic}: we observe fluctuations in fairness metrics at intermediate privacy levels, revealing a more complex, non-linear relationship between anonymity strength and fairness outcomes. A similar trend is observed in Figure~\ref{fig:main_results_adult_race} for the \adult{} dataset with \race{} as the protected attribute, where both SPD and EOD exhibit slight increases for some values of $k$, but not in a strictly increasing manner. Furthermore, as shown in Figures~\ref{fig:main_results_compas_sex}--\ref{fig:main_results_acsincome_race}, at least two group fairness metrics degrade across the \compas{} and \acsincome{} datasets, regardless of whether \gender{} or \race{} is used as the protected attribute. These findings highlight the broader implications of anonymization on fairness metrics across different datasets and protected attributes.}

\paragraph{Individual fairness.} Figure~\ref{fig:main_results_adult_gender} also reveals diverse trends across the three evaluated metrics. ALF shows \textit{non-monotonic} and \textit{method-dependent} behavior. Under \kanon{} (and $0.55$-closeness), ALF generally worsens for $2 \leq k \leq 50$, indicating that stricter anonymity levels increase model sensitivity to small input changes, suggesting reduced robustness in preserving Lipschitz fairness. In contrast, under \ldiv{} and $0.45$-closeness, ALF improves, highlighting their potential to better preserve prediction stability among similar individuals. A similar pattern is observed in Figure~\ref{fig:main_results_adult_race} for the \adult{} dataset with \race{} as the protected attribute.

In contrast, both ASF and NCF consistently improve across all three anonymization methods, datasets, and protected attributes (\ie, see Figure~\ref{fig:main_results_adult_gender} and Figures~\ref{fig:main_results_adult_race}--\ref{fig:main_results_acsincome_race}). Specifically, ASF shows substantial gains with increasing levels of anonymization, reflecting reduced prediction variability among similar individuals and promoting more equitable treatment at the individual level. For $k \geq 10$, ASF values stabilize near zero, indicating high alignment in predictions across local neighborhoods. NCF also improves with anonymization, though more gradually, as it relies on exact prediction matches. For instance, \kanon{} achieves a sharp drop in NCF at lower $k$-values, with smaller gains at higher levels. These trends reflect the inherent differences between ASF and NCF: while ASF penalizes even small deviations, NCF uses a binary notion of consistency (see Eq.~\eqref{eq:ncf}). Overall, the smoothing effect introduced by generalization appears to benefit individual fairness, consistent with similar effects observed under differential privacy~\cite{Dwork2012}.

\paragraph{Utility.} While privacy-preserving transformations reduce the risk of re-identification, they also degrade model utility~\cite{Brickell2008}, consistent with the well-known privacy-utility trade-off. Metrics such as accuracy, F1-score, and ROC AUC consistently decline as anonymization parameters are tightened. Notably, the lowest utility results are observed at $k=2$, as \ldiv{} and \tclos{} enforce the presence of at least two distinct classes within equivalence classes, leading to reduced predictive performance. However, utility metrics begin to recover as $k$ increases, as there might have a majority class decision even under \ldiv{} or \tclos. Despite this recovery, utility metrics remain below the baseline ($k=1$) and below the levels achieved under \kanon{}, underscoring the persistent trade-offs between privacy and utility.

\subsection{Detailed Analysis \#1: Record Suppression Levels and Their Effect on Fairness in ML} \label{sub:results_suppression}

Following the findings of Section~\ref{sub:main_results}, to answer \textbf{RQ2} from Section~\ref{sub:research_questions}, we now investigate the impact of the allowed record suppression level (\texttt{supp\_level}) on fairness and utility in ML. Record suppression, which involves removing rows during anonymization, can exclude outliers from the dataset, potentially worsening fairness metrics as certain demographic groups may be disproportionately impacted. To understand this effect, we vary the suppression level ($\texttt{supp\_level} \in \{10, 20, 30, 40, 50\}$) and evaluate its impact on both group and individual fairness metrics, as well as utility.
For the remainder of this analysis, the anonymity parameters are fixed at $k=10$ for \kanon{} and $t=0.5$ for \tclos.
The choice of $k=10$ represents a moderate level of anonymity, balancing privacy protection and data utility. Similarly, $t=0.5$ ensures a reasonable level of distributional closeness under \tclos{}, reflecting practical settings often used in real-world applications. Fixing these parameters allows for a focused evaluation of how varying suppression levels impact fairness and utility, independent of additional variability introduced by the anonymity parameters.

\begin{figure*}[!htb]
    \centering
    \includegraphics[width=0.84\linewidth]{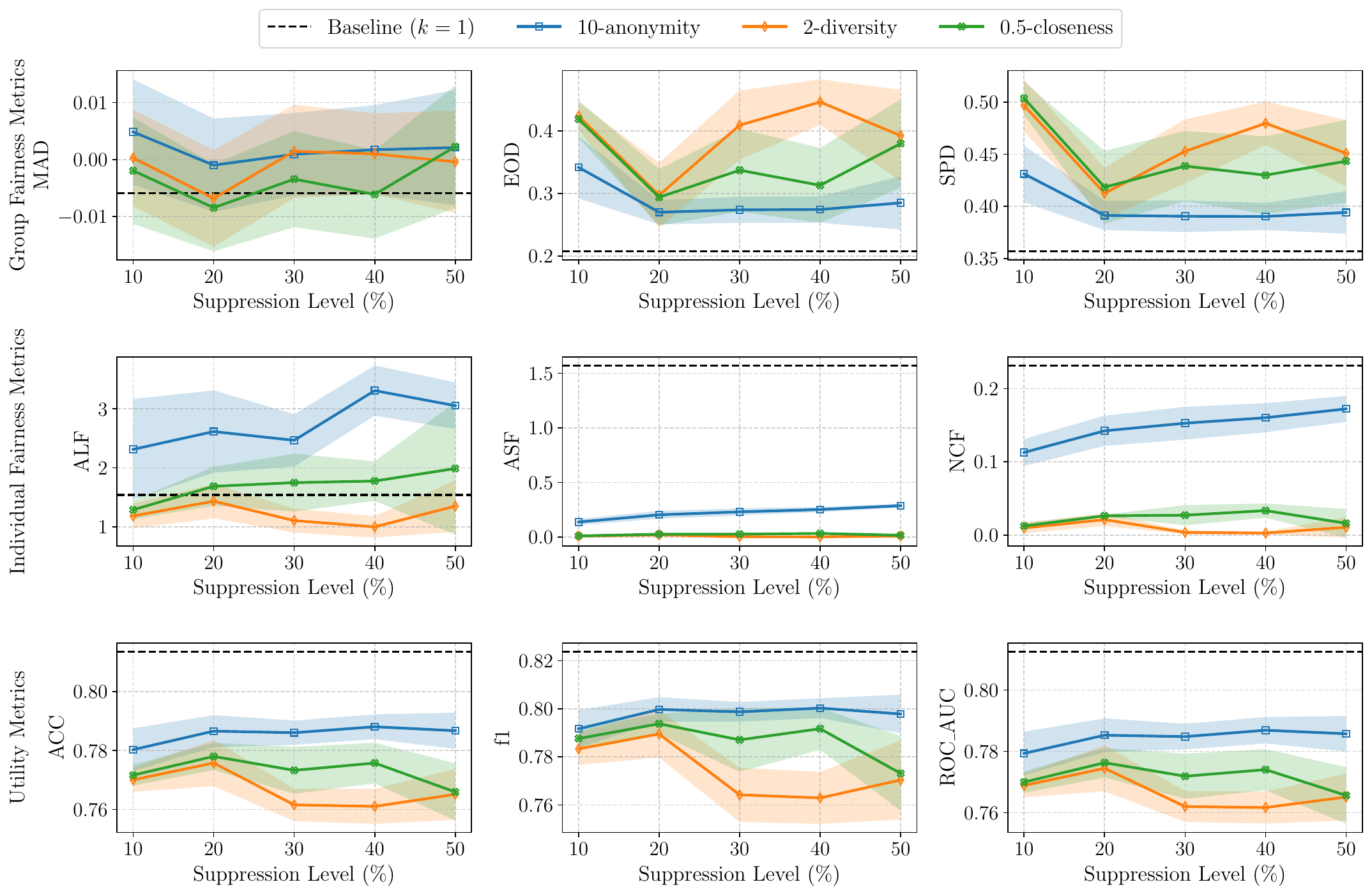}
    \caption{Effect of allowed record suppression level ($\texttt{supp\_level} \in \{10, 20, 30, 40, 50\}$) in anonymization techniques ($10$-anonymity, $2$-diversity, $0.5$-closeness) on group fairness (MAD, EOD, SPD), individual fairness (ALF, ASF, NCF), and utility (Accuracy, F1-score, ROC AUC) metrics in ML. The results are derived from the \adult{} dataset, using \gender{} as the protected attribute for fairness evaluation.}
    \label{fig:suppression_results_adult_gender}
\end{figure*}

From the results for all datasets and protected attributes presented in Figure~\ref{fig:suppression_results_adult_gender} and Figures~\ref{fig:suppression_results_adult_race}--\ref{fig:suppression_results_acsincome_race}, the impact of suppression levels on fairness and utility metrics reveals nuanced patterns. 
For instance, the effect of suppression on group fairness metrics such as SPD, EOD, and MAD is more mixed. In some cases, increasing suppression levels slightly improve these metrics, likely due to the exclusion of outliers that disproportionately skew fairness. For example, SPD and EOD in the \adult{} dataset with \race{} as the protected attribute (Figure~\ref{fig:suppression_results_adult_race}) exhibit slight improvements at moderate suppression levels. However, in other cases, particularly for the \acsincome{} dataset (Figures~\ref{fig:suppression_results_acsincome_sex} and~\ref{fig:suppression_results_acsincome_race}), the metrics remain stable or even worsen with higher suppression levels, suggesting that the removal of data points introduces disparities between demographic groups.

In addition, as the suppression level increases, individual fairness metrics such as ALF, ASF, and NCF show some degradation across all anonymization techniques and datasets. Higher suppression levels remove more rows, which disproportionately affects local neighborhoods and similar instances, introducing inconsistencies in predictions. For instance, in the \adult{} dataset with \gender{} as the protected attribute (Figure~\ref{fig:suppression_results_adult_gender}), NCF values increase steadily as the suppression level rises, signaling worsening prediction consistency. Similarly, both ALF and ASF metrics are also worsened, highlighting increased variations in predictions for similar inputs. This trend suggests that higher suppression levels may hinder the ability of anonymization methods to preserve fairness at the individual level.

In terms of utility, record suppression shows mixed impacts on utility metrics such as ACC, F1-score, and ROC AUC. In most cases, utility metrics remain stable or even improve slightly as the suppression level increases, likely due to the removal of noisy or less representative data. However, in a few instances (for \ldiv{} and \tclos), utility decreases, particularly at higher suppression levels, reflecting the trade-off between privacy and maintaining a dataset that is representative of the original population. These observations suggest that while suppression can enhance privacy, \textbf{\textit{its effect on utility is nuanced and may vary depending on the dataset and the anonymization method applied.}}

\subsection{Detailed Analysis \#2: ML Fairness Across Target Distribution Variations} \label{sub:results_target_distribution}

\begin{figure*}[!htb]
    \centering
    \includegraphics[width=0.84\linewidth]{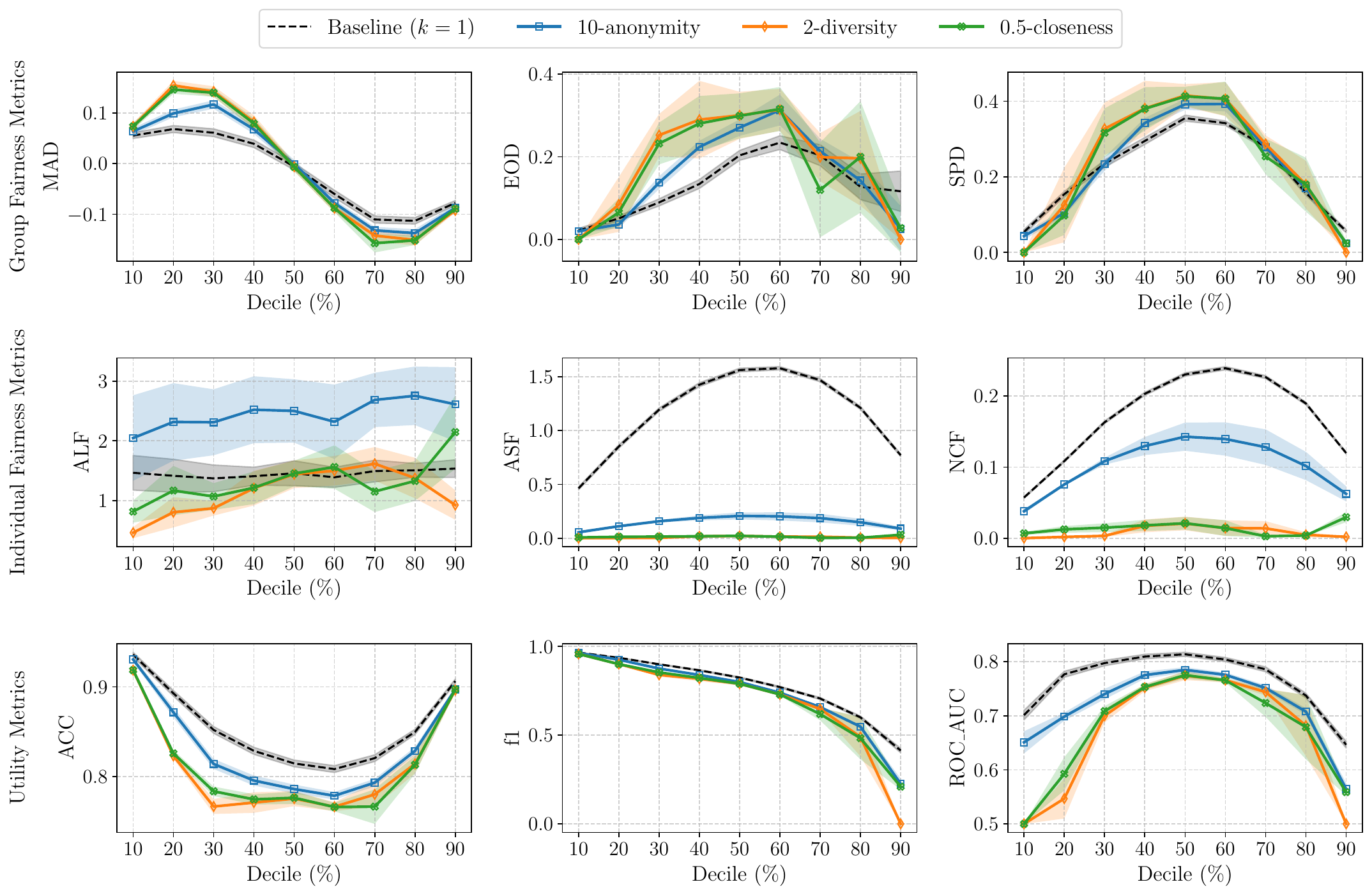}
    \caption{Effect of target distribution variation (\ie, thresholded at deciles ranging from 10\% to 90\%) on anonymity techniques ($10$-anonymity, $2$-diversity, $0.5$-closeness) regarding group fairness (MAD, EOD, SPD), individual fairness (ALF, ASF, NCF), and utility (Accuracy, F1-score, ROC AUC) metrics in ML. Results are presented for the \adult{} dataset, with \gender{} serving as the protected attribute for fairness evaluation.}    
    \label{fig:target_distribution_results_adult_gender}
\end{figure*}

To answer \textbf{RQ3} from Section~\ref{sub:research_questions}, we now investigate the impact of target distribution variations on fairness and utility in ML models. {With both \adult{} and \acsincome{} datasets, we modify the distribution of the \texttt{income} target variable by thresholding it at deciles ranging from 10\% to 90\%, simulating shifts in the balance between positive and negative classes. Similarly, for the \compas{} dataset, we modify the distribution of the \texttt{COMPAS risk score} target variable by thresholding it from scores 1 to 9.}
Figure~\ref{fig:target_distribution_results_adult_gender} presents the results of these experiments focusing on the \adult{} dataset with \gender{} as the protected attribute. 
Additional results for the \adult{} dataset with \race{}, and both \gender{} and \race{} in the \acsincome{} and \compas{} datasets are shown in Figures~\ref{fig:target_distribution_results_adult_race}--\ref{fig:target_distribution_results_ACSIncome_race}.

The results across all datasets and protected attributes (Figure~\ref{fig:target_distribution_results_adult_gender} and Figures~\ref{fig:target_distribution_results_adult_race}--\ref{fig:target_distribution_results_ACSIncome_race}) reveal consistent trends regarding the impact of anonymization techniques on fairness and utility metrics. 
Notably, the anonymized models tend to follow the same general shape as the baseline across varying deciles, indicating that the overall distributional patterns of fairness and utility metrics are preserved under anonymization. 
However, the magnitude of these metrics differs: anonymization consistently worsens group fairness metrics such as MAD, SPD, and EOD, reflecting an amplification of disparities between demographic groups. 
Conversely, individual fairness metrics, particularly ASF and NCF, generally benefit from anonymization, showing improved consistency in predictions for similar inputs. 
On the other hand, utility metrics such as ACC, F1-score, and ROC AUC are negatively impacted, with performance often falling below the baseline. 
These findings align with the conclusions drawn for \textbf{RQ1} in Section~\ref{sub:main_results}, further underscoring the trade-offs between privacy, fairness, and utility introduced by anonymization techniques.

\subsection{Detailed Analysis \#3: Impact of Data Size on Fairness in ML} \label{sub:results_data_size}

\begin{figure*}[!htb]
    \centering
    \includegraphics[width=0.84\linewidth]{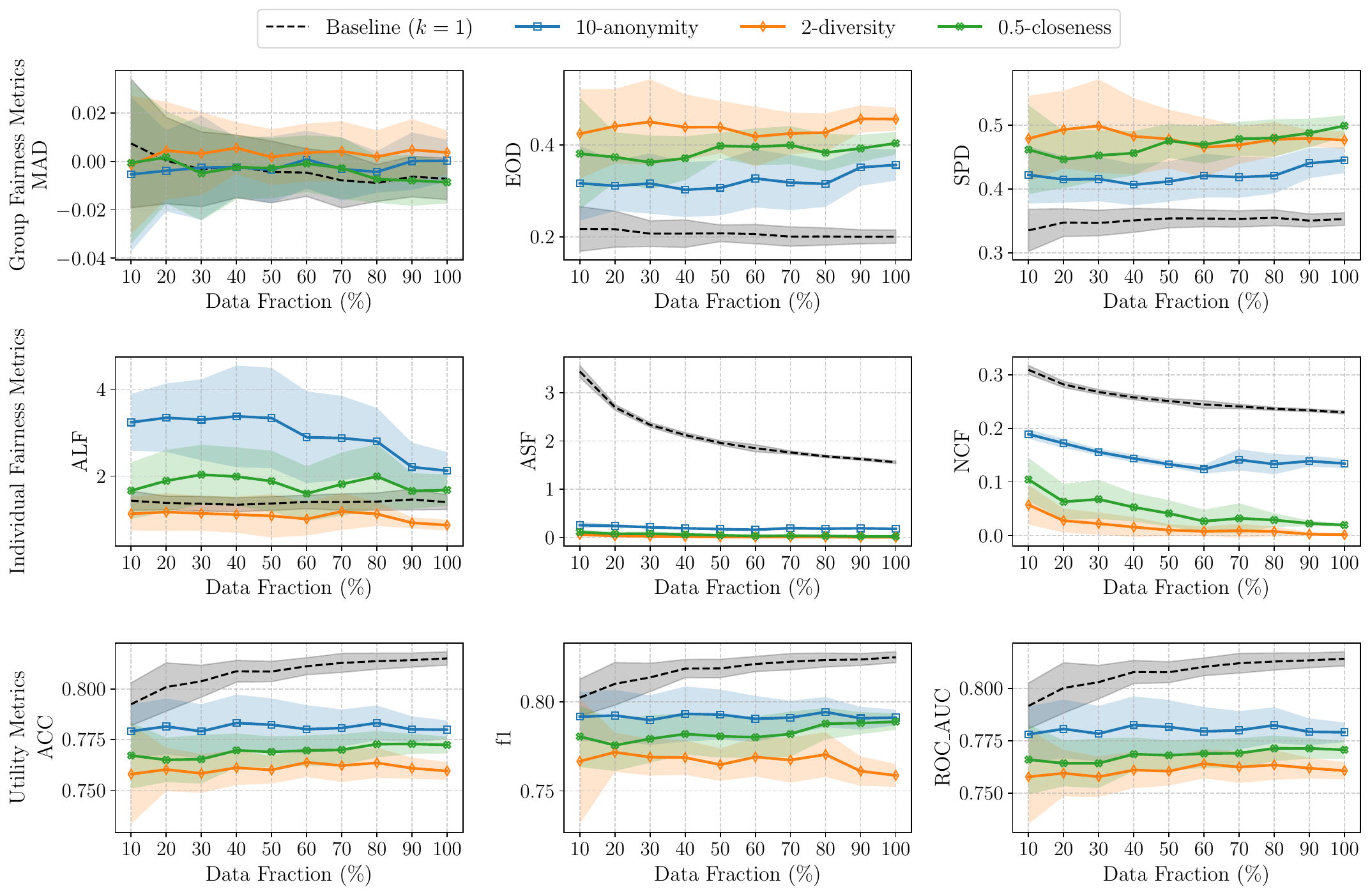}
    \caption{Effect of varying data fraction on the performance of anonymity techniques ($10$-anonymity, $2$-diversity, $0.5$-closeness) in terms of group fairness metrics (MAD, EOD, SPD), individual fairness metrics (ALF, ASF, NCF), and utility metrics (Accuracy, F1-score, ROC AUC) in ML. This analysis is performed using the \adult{} dataset, considering \gender{} as the protected attribute for fairness evaluation.}
\label{fig:data_fraction_results_adult_gender}
\end{figure*}

To address whether data fraction impacts fairness as formulated in \textbf{RQ4} in Section~\ref{sub:research_questions}, we analyze the behavior of anonymization techniques across subsampled data fractions. Specifically, we vary the data fraction from 10\% to 100\% of the original dataset, subsampling the data randomly at each fraction level. Figure~\ref{fig:data_fraction_results_adult_gender} illustrates the results for the \adult{} dataset with \gender{} as the protected attribute; additional results for other datasets and protected attributes are provided in Figures 21--25 in Appendix~\ref{app:add_results}. 

Figure~\ref{fig:data_fraction_results_adult_gender} and Figures~\ref{fig:data_fraction_results_adult_race}--\ref{fig:data_fraction_results_ACSIncome_race} reveal that performance and fairness metrics under anonymization techniques follow a relatively stable pattern across different data fraction levels sampled from 10\% to 100\%. 
For instance, anonymization introduces consistent increases in group fairness disparities compared to the baseline, regardless of the dataset size. 
Similarly, we observe a positive effect of anonymization on individual fairness that remains stable across data sizes.
Performance metrics improve gradually with increasing data, as expected.
In conclusion, these findings suggest that the trade-offs between privacy, fairness, and utility are primarily driven by the anonymization techniques and their parameter settings, rather than by the scale of the dataset. 
The random sampling of fractions introduces minor variability but does not fundamentally alter the patterns observed.

\subsection{Detailed Analysis \#4: Comparison of ML Classifiers in Fairness Under Anonymization} \label{sub:results_ml_classifier}

\begin{figure*}[!htb]
    \centering
    \includegraphics[width=0.84\linewidth]{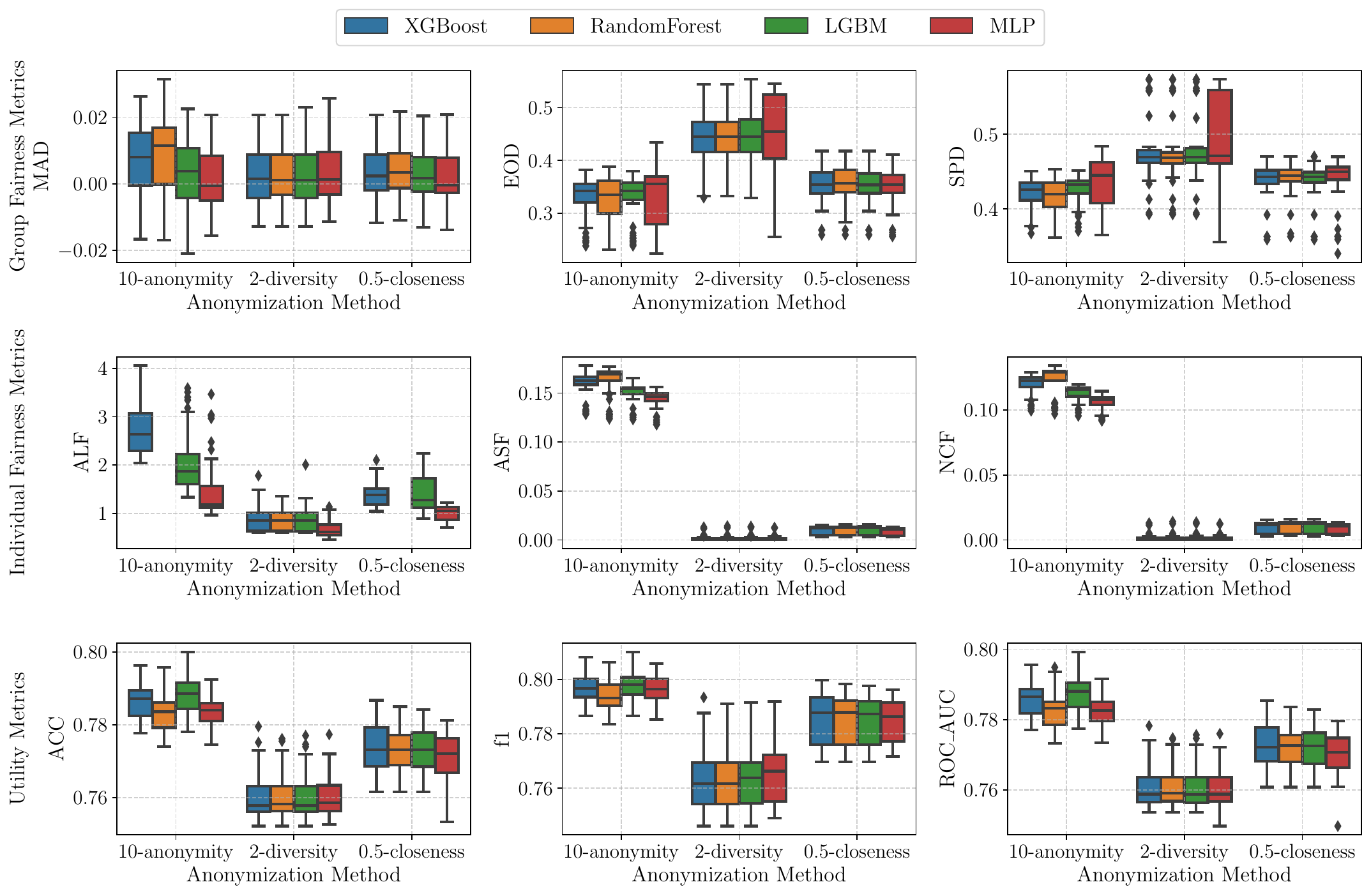}
    \caption{Comparison of the impact of different state-of-the-art ML classifiers on anonymized dataset (\kanon, \ldiv, \tclos) and relation to group fairness (MAD, EOD, SPD), individual fairness (ALF, ASF, NCF), and utility (Accuracy, F1-score, ROC AUC) metrics in ML. Results are based on the \adult{} dataset, with \gender{} as the protected attribute for fairness evaluation.}
    \label{fig:classifier_results_adult_gender}
\end{figure*}

Finally, to address \textbf{RQ5} in Section~\ref{sub:research_questions}, we compare the impact of anonymization techniques on fairness and utility across multiple state-of-the-art ML classifiers, including LGBM~\cite{LGBM}, Random Forest~\cite{Breiman2001}, Neural Networks (MLP), and XGBoost~\cite{XGBoost}. Figure~\ref{fig:classifier_results_adult_gender} illustrates the results using the \adult{} dataset with \gender{} as the protected attribute; additional comparisons are shown in Figures~\ref{fig:classifier_results_adult_race}--\ref{fig:classifier_results_ACSIncome_race}.

From Figure~\ref{fig:classifier_results_adult_gender} and Figures~\ref{fig:classifier_results_adult_race}--\ref{fig:classifier_results_ACSIncome_race}, results show that the trends observed with XGBoost in previous sections (\textbf{RQ1}–\textbf{RQ4}) are consistent across other classifiers. 
Specifically, anonymization continues to negatively affect group fairness and to improve individual fairness.
Moreover, as expected, utility degrades under anonymization, regardless of the classifier. 
However, the magnitude of degradation varies slightly. 
XGBoost and LGBM tend to retain higher performance than Random Forest and MLP, especially in terms of F1-score and ROC AUC.
{Although minor differences exist, such as XGBoost's marginal advantage in utility and fairness stability, the median values for all fairness and utility metrics are nearly identical across classifiers. This visual and statistical consistency reinforces that our findings are not specific to XGBoost. Hence, the conclusions drawn from XGBoost experiments can be considered broadly representative, supporting the generalizability of our results across different learning algorithms.}

\subsection{General Findings and Practical Guidelines} \label{sub:guidelines}

This section summarizes the key findings from Sections~\ref{sub:main_results}--\ref{sub:results_ml_classifier} and provides actionable guidelines for practitioners.

\textcolor{black}{
\begin{leftbar}
\noindent \textbf{Fairness \emph{vs.} anonymization has \textcolor{red}{non-monotonic} behavior:} 
While one might expect fairness to degrade monotonically with stricter privacy levels (\ie, higher $k$, $\ell$, or $t$), our findings consistently show non-linear behavior across datasets and metrics. This is due to the complex interplay between generalization, suppression, and model learning dynamics. Specifically, anonymization alters the statistical dependencies between features (\ie, both sensitive and non-sensitive), as well as between features and the target. Since generalization may be applied to any quasi-identifier, including the sensitive attribute, the relative influence of the sensitive feature on the model can increase or decrease in unpredictable ways compared to the non-anonymized case ($k = 1$). These shifts can result in fluctuating fairness outcomes that do not follow a simple degradation trend.
\end{leftbar}}

\begin{leftbar}
\noindent \textbf{Anonymity \textcolor{red}{negatively} impacts group fairness in ML:}  
{Anonymization methods generally degraded group fairness metrics (\ie, MAD, EOD, and SPD) as privacy parameters tightened. By coarsening quasi‑identifiers and suppressing records, these anonymity-based methods distort the true joint distribution of $(A,X,Y)$, skew subgroup prevalences and error rates, and thereby amplify accuracy and positive‑rate gaps between privileged and unprivileged groups.}
\end{leftbar}

\begin{leftbar}
\noindent \textbf{Anonymity \textcolor{blue}{positively} impacts individual fairness in ML:}  
{Anonymization techniques improve similarity-based individual fairness metrics such as ASF and NCF by inducing input homogeneity through generalization, which leads to more consistent predictions for similar individuals. A similar effect has been observed with differential privacy mechanisms~\cite{Dwork2012}. However, since these improvements emerge from changes in data structure rather than targeted fairness interventions, they should be interpreted carefully to avoid overclaiming fairness benefits (see~\cite{Aivodji2019,Aivodji2021,meding2024fairness}).}
\end{leftbar}

\begin{leftbar}
\noindent \textbf{Higher record suppression levels \textcolor{red}{negatively} impacts individual fairness in ML:} Our findings reveal that higher suppression thresholds decrease the density of local equivalence classes, undermining the model’s capacity to generate consistent predictions for nearby instances. In contrast, group fairness metrics show mixed behavior, with minimal or no consistent degradation across suppression levels. 
\end{leftbar}

\begin{leftbar} \noindent \textbf{Target distribution variations \textcolor{red}{amplify} group fairness disparities but \textcolor{blue}{stabilize} individual fairness:}
Adjusting the distribution of the target variable (\eg, varying thresholds for binarization) has a significant impact on group fairness metrics, with disparities (\eg, EOD and SPD) peaking at middle deciles. However, individual fairness metrics such as ASF and NCF remain relatively stable across target variations, benefiting from anonymization-induced generalization. 
\end{leftbar}

\begin{leftbar} \noindent \textbf{Data size variations have minimal impact on fairness trends:}
\noindent Across subsampled data fractions (10\% to 100\%), fairness and utility metrics follow consistent trends. Group fairness metrics remain negatively impacted by anonymization, while individual fairness metrics improve consistently. Utility metrics exhibit minor degradation at smaller fractions but generally remain stable. 
\end{leftbar}

\begin{leftbar} \noindent \textbf{XGBoost findings generalize across ML classifiers:}
\noindent Fairness and utility trends observed with XGBoost are consistent across other classifiers, such as Random Forest and Neural Networks (MLP). While XGBoost often achieves slightly better utility, the broader patterns, \ie, negative group fairness impacts, positive individual fairness outcomes, and utility trade-offs, remain similar. This consistency supports the generalizability of conclusions drawn from XGBoost experiments. 
\end{leftbar}

{
\textbf{Guidelines.} Based on our empirical findings, we propose the following recommendations for practitioners working with anonymized datasets in ML pipelines:

\begin{itemize}
    \item \textbf{Use moderate privacy parameters} (\eg, $k=10$, $t=0.5$) to balance privacy, fairness, and utility. Stricter settings can severely degrade group fairness and predictive performance..

    \item \textbf{Handle record suppression carefully} as high suppression thresholds may disproportionately remove minority or outlier samples, harming both individual and group fairness. When possible, combine suppression with imputation or domain-specific filtering to minimize bias.

    \item \textbf{Avoid median splits when binarizing continuous target variable}, as they tend to maximize group disparities. Evaluate multiple cutpoints across deciles. We found that thresholds below the $30$th or above the $70$th percentile yields lower group disparities; whereas median splits ($40$--$60$\%) tend to exacerbate them. 

    \item \textbf{Interpret individual fairness improvements cautiously.} Improvements in individual‐fairness scores (ASF, NCF) are primarily due to feature homogenization and may not reflect  \textit{``genuine improvements in equitable treatment''}. Practitioners should therefore exercise caution and perform targeted case audits to avoid the risk of ``fair washing/hacking''~\cite{Aivodji2019,Aivodji2021,meding2024fairness}, for example.
\end{itemize}
}

\section{Conclusion and Perspectives} \label{sec:conclusion}

This study systematically audits the tradeoff between anonymization techniques and fairness in ML. 
Through a comprehensive analysis, we evaluated the effects of three well-known anonymization methods (\kanon, \ldiv, and \tclos) on group and individual fairness metrics, as well as utility metrics, across multiple datasets and ML classifiers. 
By addressing five key research questions, our findings highlight the inherent trade-offs between privacy, fairness, and utility in anonymized ML models.
{Overall, our results show that anonymization tends to negatively affect group fairness metrics, often exacerbating disparities between demographic groups as privacy constraints increase. 
In contrast, similarity-based individual fairness metrics tend to improve under stronger anonymization, driven by increased data homogeneity. 
While this effect aligns with phenomena observed in differential privacy~\cite{Dwork2012}, it stems from data structural smoothing rather than fairness-aware optimization, and must therefore be interpreted with caution.
Importantly, we demonstrate that these findings are robust across different datasets, protected attributes, target distributions, data scales, suppression levels, and learning algorithms, confirming the generalizability of our conclusions.
This work opens several promising research directions. 
A theoretical characterization of how anonymization influences fairness, \eg, through the lenses of causal modeling or information-theoretic frameworks, remains an open challenge.
\textcolor{black}{In particular, understanding the impact on individual fairness, where local input neighborhoods are disrupted by generalization and suppression, calls for a principled framework to describe how anonymization transforms the geometry of the feature space.}
Future efforts could also focus on designing privacy mechanisms that jointly optimize for fairness constraints, or on adapting anonymization strategies to richer settings such as multi-class classification or regression.
}


\begin{ack}
This work has been partially supported by the French National Research Agency (ANR), under contracts ``ANR-24-CE23-6239'' JCJC project AI-PULSE, ``ANR-19-P3IA-0003'' MIAI @ Grenoble Alpes, ``ANR-22-CE39-0002'' JCJC project EQUIHID, and ``ANR 22-PECY-0002'' IPOP (Interdisciplinary Project on Privacy) project of the Cybersecurity PEPR.
\end{ack}


\bibliography{ref}

\appendix
\onecolumn

\section{Description of Datasets} \label{app:datasets}

\begin{itemize}
    \item \textbf{Adult Dataset:} The \adult{} dataset, obtained from the UCI ML repository~\cite{ADULT} originates from the 1994 U.S. Census Bureau database. For this study, we use the \texttt{Reconstructed Adult} dataset as described in~\cite{ding2021retiring}, where the target variable, \textit{income}, is represented as a discrete value. This variable can be binarized into $> \mathsf{ThresholdIncome}$ or $\leq \mathsf{ThresholdIncome}$, where $\mathsf{ThresholdIncome}$ serves as a configurable threshold. By default, the threshold is set to the median income value; however, in Section~\ref{sub:results_target_distribution}, we vary this threshold to analyze the impact of target distribution shifts on fairness and utility. After preprocessing and data cleaning, the dataset consists of $n=45,849$ samples and $10$ discrete and categorical attributes. The \textit{protected attributes} used for fairness evaluation are gender and race, while the \textit{target attribute} is income. 

    \item \textbf{Compas Dataset:} The \compas{} dataset, curated by ProPublica~\cite{compas}, contains defendants from Broward County, Florida, screened between 2013 and 2014. We restrict to Black and White individuals who received a COMPAS risk score within 30 days of arrest. After preprocessing, the dataset comprises $n=5,278$ and five attributes: \texttt{race}, \texttt{sex}, \texttt{age}, \texttt{priors\_count}, and \texttt{days\_b\_screening\_arrest}. The target variable is the COMPAS risk score (\texttt{v\_decile\_score}), which consists of a rating of $1-10$; the higher the score, the more likely the defendant is to re-offend. This target variable can also be binarized into $> \mathsf{RiskScore}$ or $\leq \mathsf{RiskScore}$, where $\mathsf{RiskScore}$ serves as a configurable threshold. By default, the threshold is set to the median score value (\ie, 3); however, as in Section~\ref{sub:results_target_distribution}, we vary this threshold (1--10) to analyze the impact of target distribution shifts on fairness and utility. The \textit{protected attributes} used for fairness evaluation are gender and race. 

    \item \textbf{ACSIncome Dataset:} The \acsincome{} dataset, sourced from the U.S. Census Bureau’s American Community Survey (ACS), represents a geographically distributed sample of individuals across U.S. states. Similar to the Adult dataset, the target variable \textit{income} is categorized as $> \mathsf{ThresholdIncome}$ or $\leq \mathsf{ThresholdIncome}$, with $\mathsf{ThresholdIncome}$ configurable to several representative income levels (median by default). 
    For this study, we use the 2018 1-Year ACS Public Use Microdata Sample (\texttt{survey\_year="2018"} and \texttt{horizon="1-Year"}) across all U.S. states.
    The whole dataset contains $n=1,599,229$ data points with 10 discrete and categorical attributes. 
    To reduce computational resource consumption, we randomly sample $10\%$ of the data. The protected attributes for \textit{fairness evaluation} are gender and race (\ie, \texttt{SEX}, \texttt{RAC1P}), while the \textit{target attribute} is income.
\end{itemize}

\section{Additional Results} \label{app:add_results}
This section presents additional analyses answering the same research questions presented in Section~\ref{sub:research_questions} for \adult{} dataset with \race{} as the protected attribute, on the \compas{} dataset with \gender{} and \race{} as protected attributes, and on the \acsincome{} dataset with \texttt{SEX} and \texttt{RAC1P} as protected attributes.

\subsection{Impact of Anonymization on Fairness in ML}

\begin{figure}[!htb]
    \centering
    \includegraphics[width=0.84\linewidth]{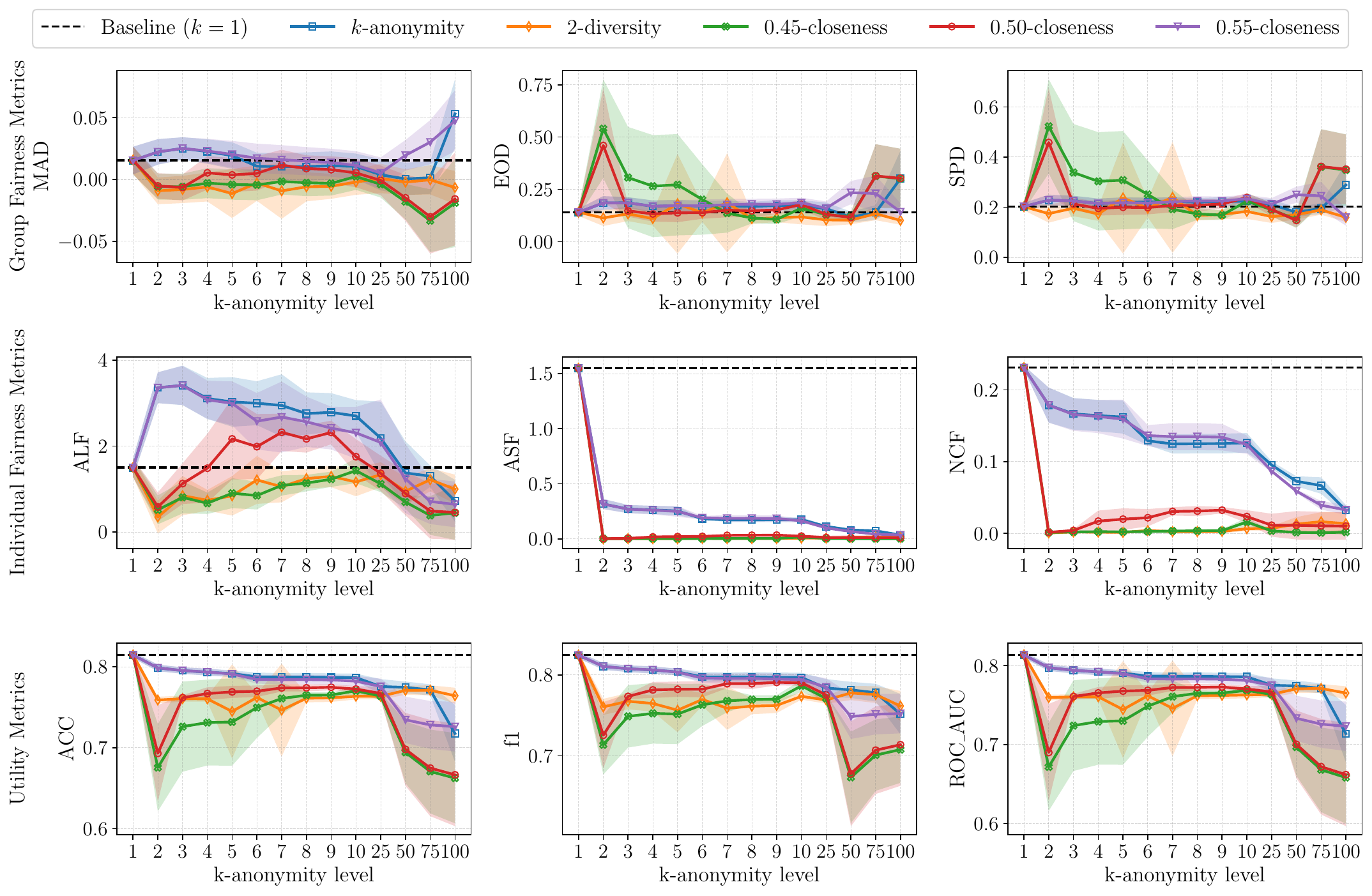}
    \caption{Impact of anonymity methods (\kanon, \ldiv, \tclos) on group fairness (MAD, EOD, SPD), individual fairness (ALF, ASF, NCF), and utility (Accuracy, F1-score, ROC AUC) metrics in ML.
    Results with the \adult{} dataset with \race{} as the protected attribute for fairness evaluation.}
    \label{fig:main_results_adult_race}
\end{figure}

\begin{figure}[!htb]
    \centering
    \includegraphics[width=0.84\linewidth]{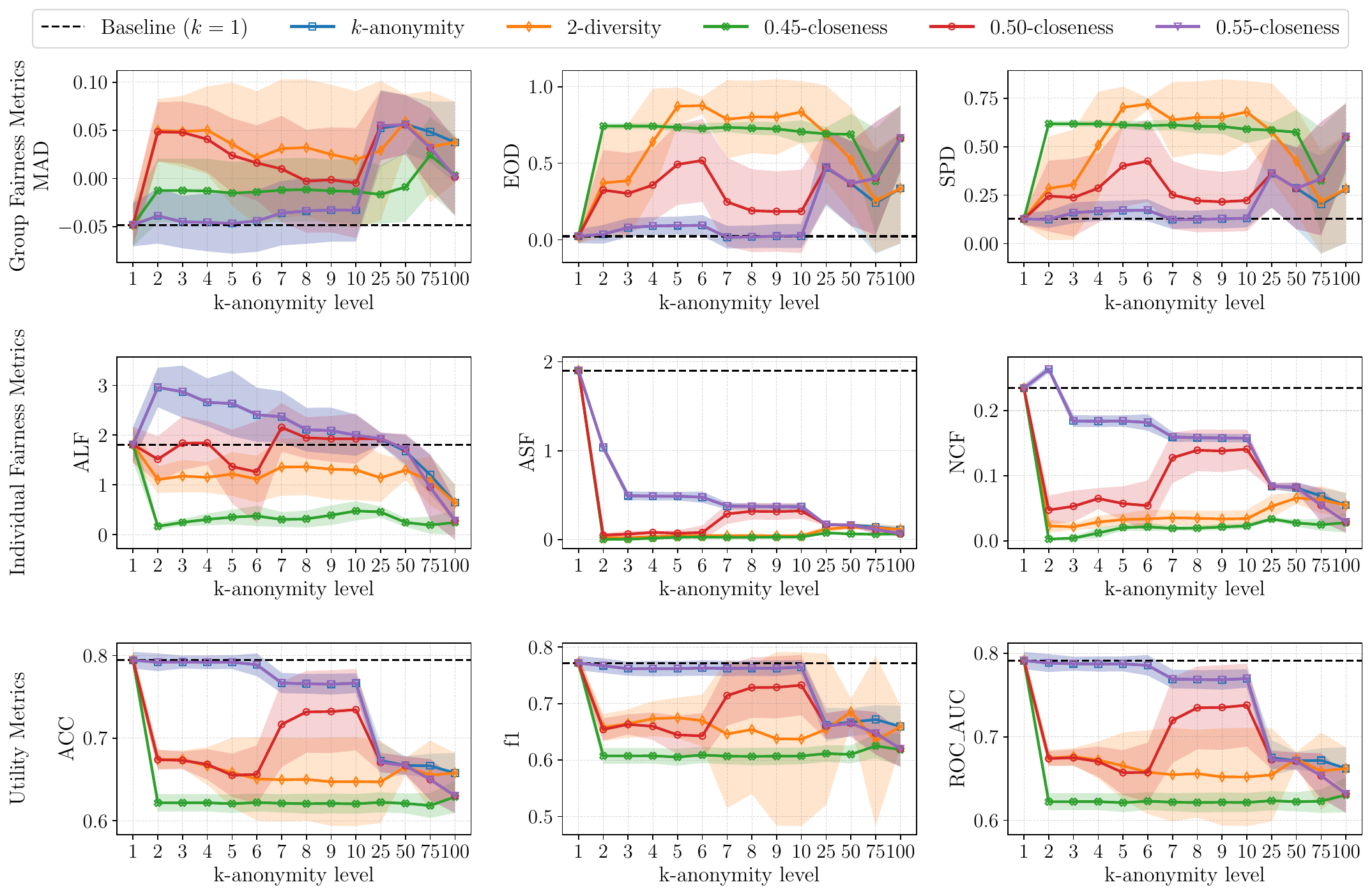}
    \caption{Impact of anonymity methods (\kanon, \ldiv, \tclos) on group fairness (MAD, EOD, SPD), individual fairness (ALF, ASF, NCF), and utility (Accuracy, F1-score, ROC AUC) metrics in ML.
    Results with the \compas{} dataset with \gender{} as the protected attribute for fairness evaluation.}
    \label{fig:main_results_compas_sex}
\end{figure}

\begin{figure}[!htb]
    \centering
    \includegraphics[width=0.84\linewidth]{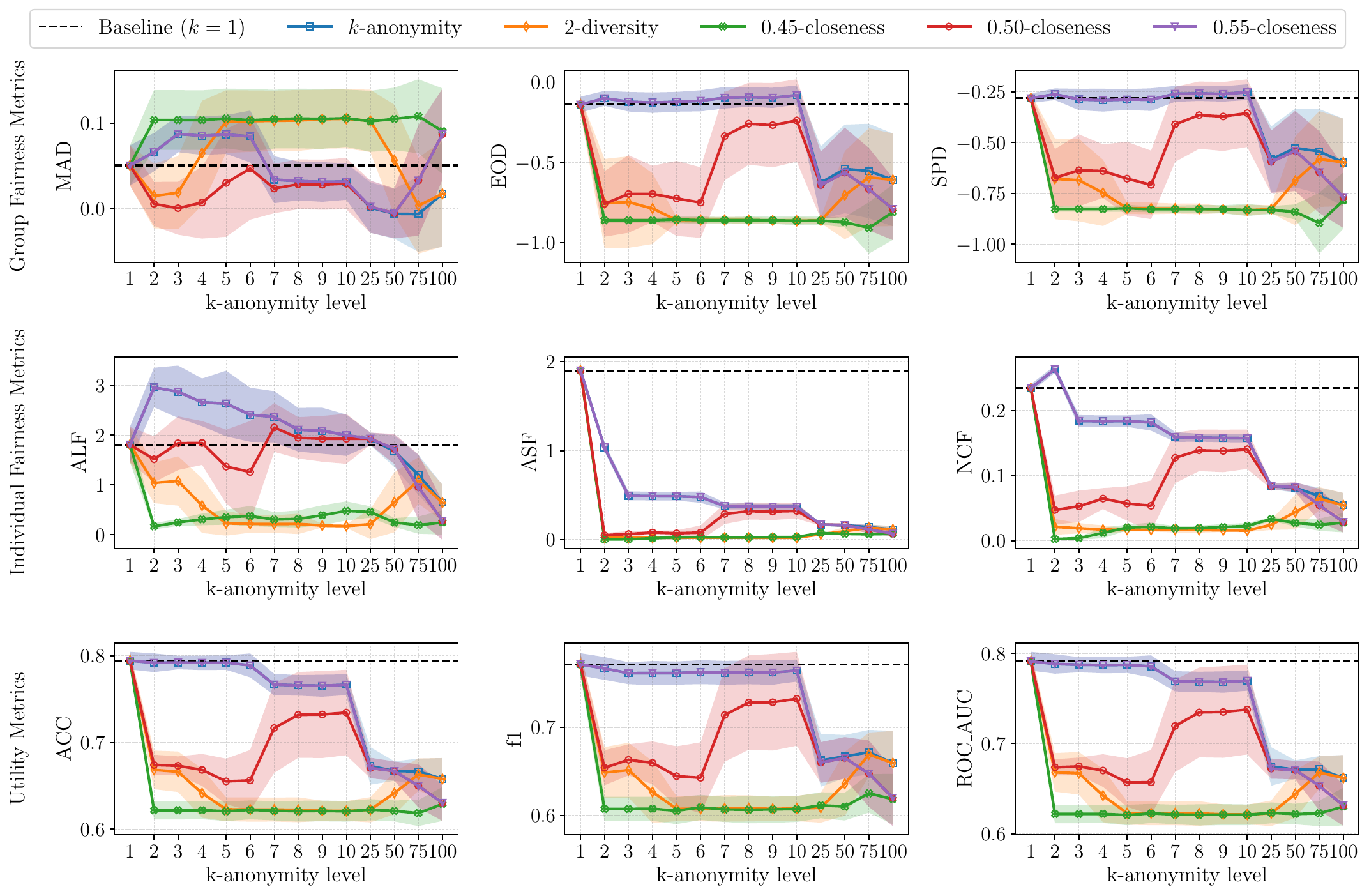}
    \caption{Impact of anonymity methods (\kanon, \ldiv, \tclos) on group fairness (MAD, EOD, SPD), individual fairness (ALF, ASF, NCF), and utility (Accuracy, F1-score, ROC AUC) metrics in ML.
    Results with the \compas{} dataset with \race{} as the protected attribute for fairness evaluation.}
    \label{fig:main_results_compas_race}
\end{figure}

\begin{figure}[!htb]
    \centering
    \includegraphics[width=0.84\linewidth]{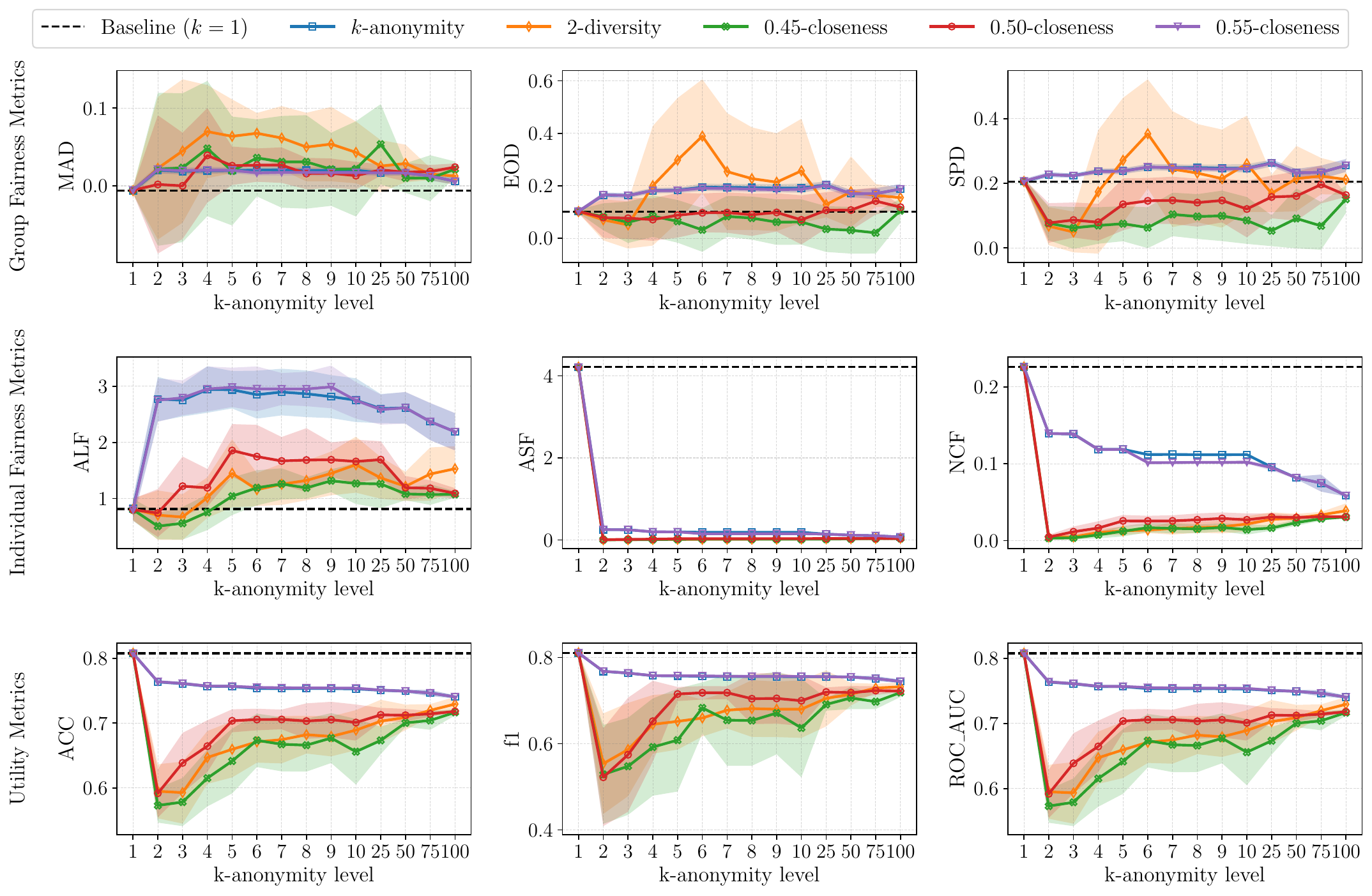}
    \caption{Impact of anonymity methods (\kanon, \ldiv, \tclos) on group fairness (MAD, EOD, SPD), individual fairness (ALF, ASF, NCF), and utility (Accuracy, F1-score, ROC AUC) metrics in ML.
    Results with the \acsincome{} dataset with \gender{} as the protected attribute for fairness evaluation.}
    \label{fig:main_results_acsincome_sex}
\end{figure}

\begin{figure}[!htb]
    \centering
    \includegraphics[width=0.84\linewidth]{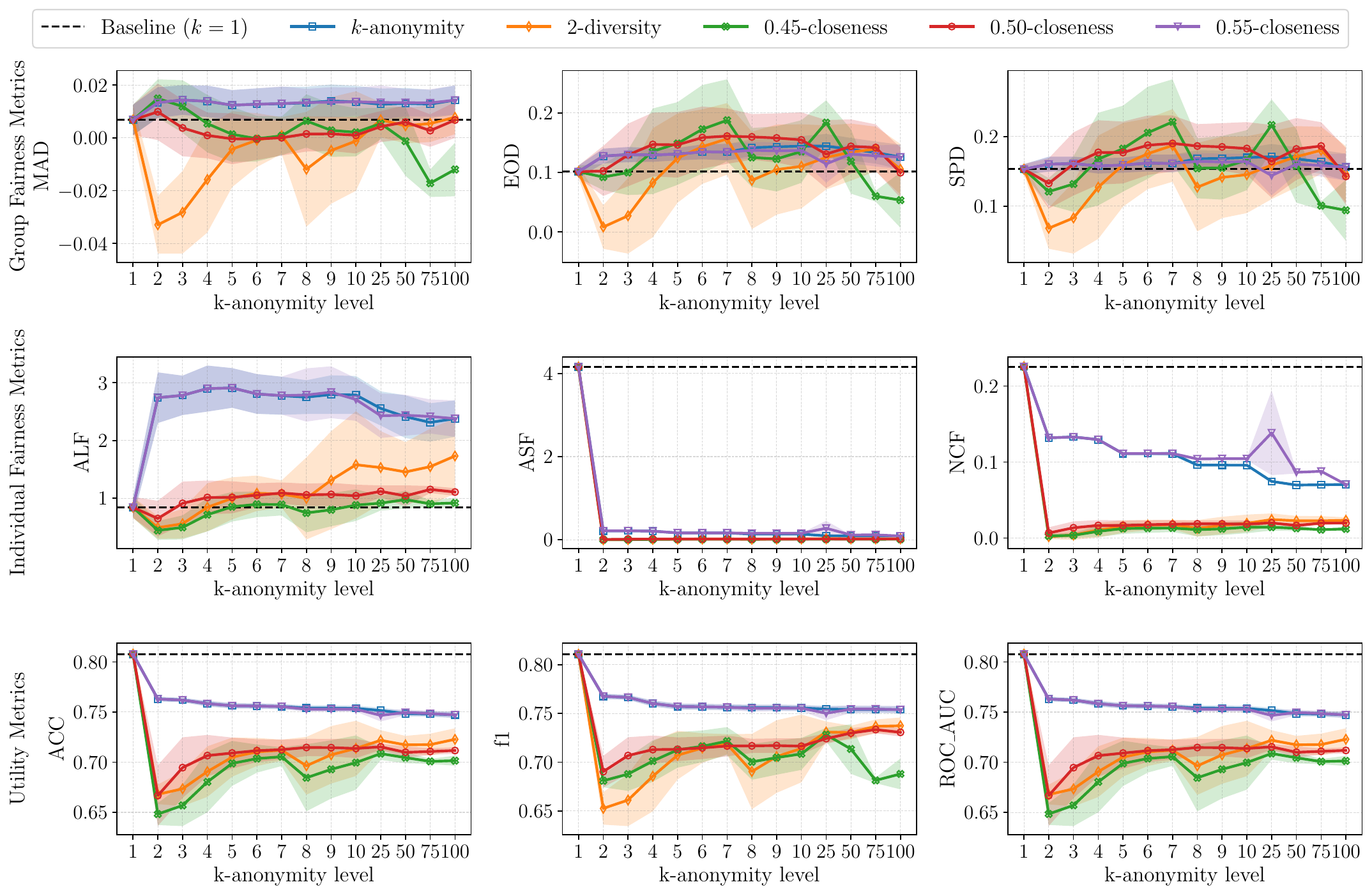}
    
    \caption{Impact of anonymity methods (\kanon, \ldiv, \tclos) on group fairness (MAD, EOD, SPD), individual fairness (ALF, ASF, NCF), and utility (Accuracy, F1-score, ROC AUC) metrics in ML.
    Results with the \acsincome{} dataset with \race{} as the protected attribute for fairness evaluation.}
    \label{fig:main_results_acsincome_race}
\end{figure}

Figures~\ref{fig:main_results_adult_race},~\ref{fig:main_results_compas_sex},~\ref{fig:main_results_compas_race},~\ref{fig:main_results_acsincome_sex}, and~\ref{fig:main_results_acsincome_race} show the impact of \kanon, \ldiv, and \tclos on group and individual fairness across different datasets. Specifically, they present results for the \adult{} dataset with \race{} as the protected attribute and for both \compas{} and \acsincome{} datasets, considering \gender{} and \race{} as protected attributes. These experiments extend the findings discussed in Section~\ref{sub:main_results}.

The results confirm that anonymization negatively affects group fairness, with the impact becoming more pronounced as privacy constraints become stricter. Conversely, while the trends in Approximate Lipschitz Fairness (ALF) vary across different settings, anonymization generally has a positive effect on individual fairness.

\subsection{Impact of Record Suppression Levels on Fairness in ML}

\begin{figure}[!htb]
    \centering
    
    \includegraphics[width=0.84\linewidth]{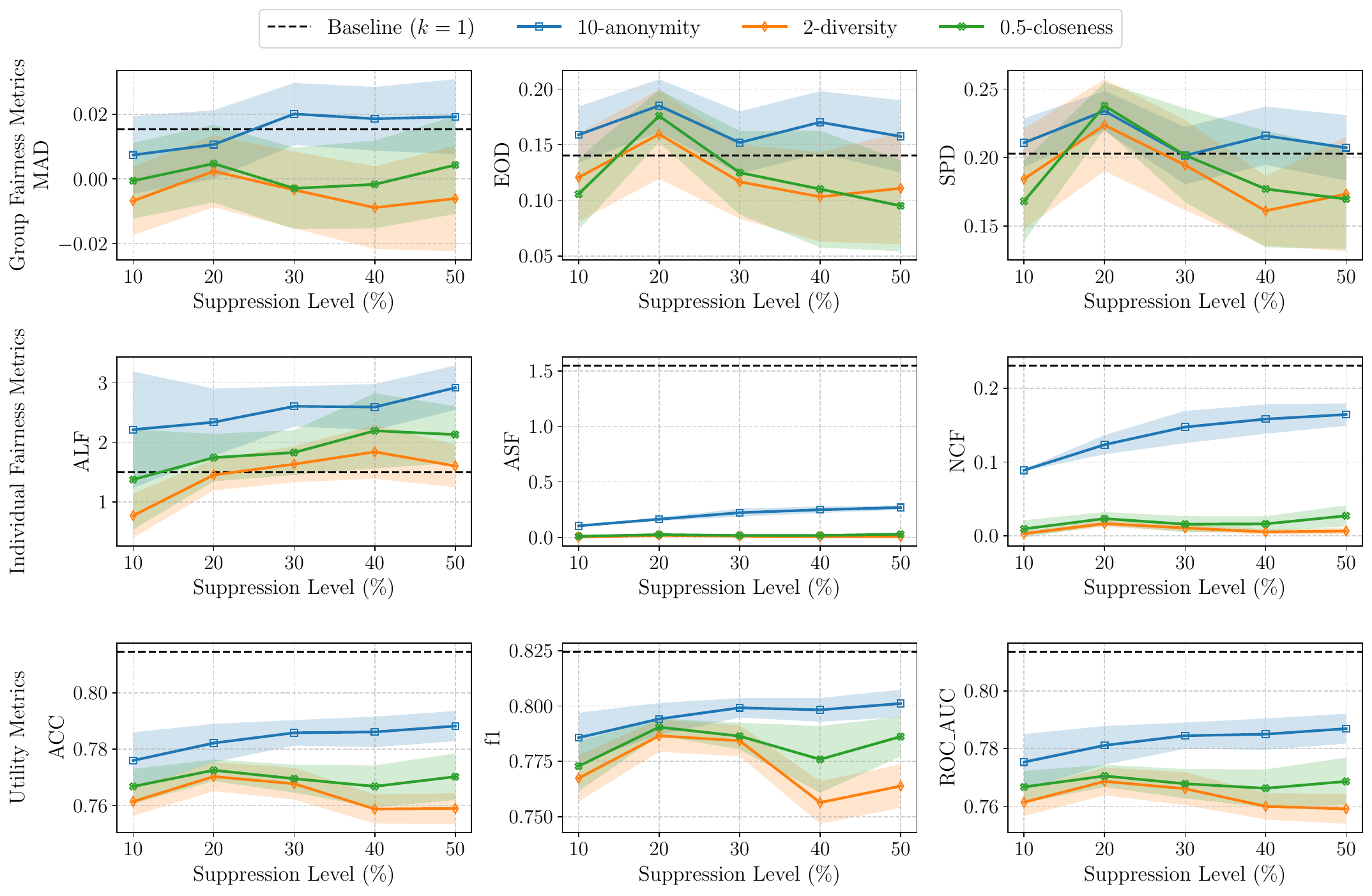}
    
    \caption{Impact of suppression level in anonymization (\kanon, \ldiv, \tclos) on group fairness (MAD, EOD, SPD), individual fairness (ALF, ASF, NCF), and utility (Accuracy, F1-score, ROC AUC) metrics in ML.
    Results with the \adult{} dataset with \race{} as the protected attribute for fairness evaluation.}
    \label{fig:suppression_results_adult_race}
\end{figure}

\begin{figure}[!htb]
    \centering
    
    \includegraphics[width=0.84\linewidth]{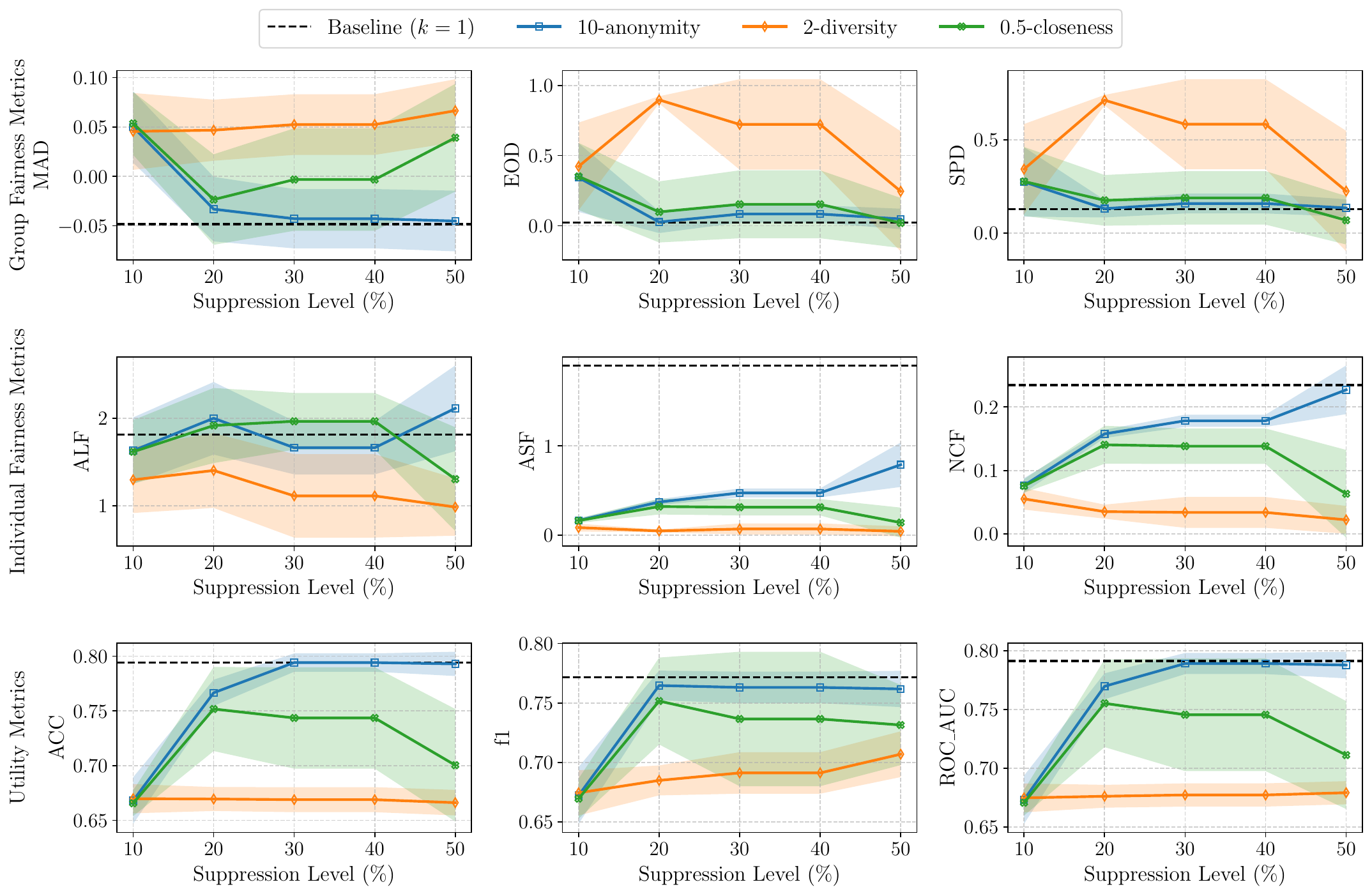}
    
    \caption{Impact of suppression level in anonymization (\kanon, \ldiv, \tclos) on group fairness (MAD, EOD, SPD), individual fairness (ALF, ASF, NCF), and utility (Accuracy, F1-score, ROC AUC) metrics in ML.
    Results with the \compas{} dataset with \gender{} as the protected attribute for fairness evaluation.}
    \label{fig:suppression_results_compas_sex}
\end{figure}

\begin{figure}[!htb]
    \centering
    
    \includegraphics[width=0.84\linewidth]{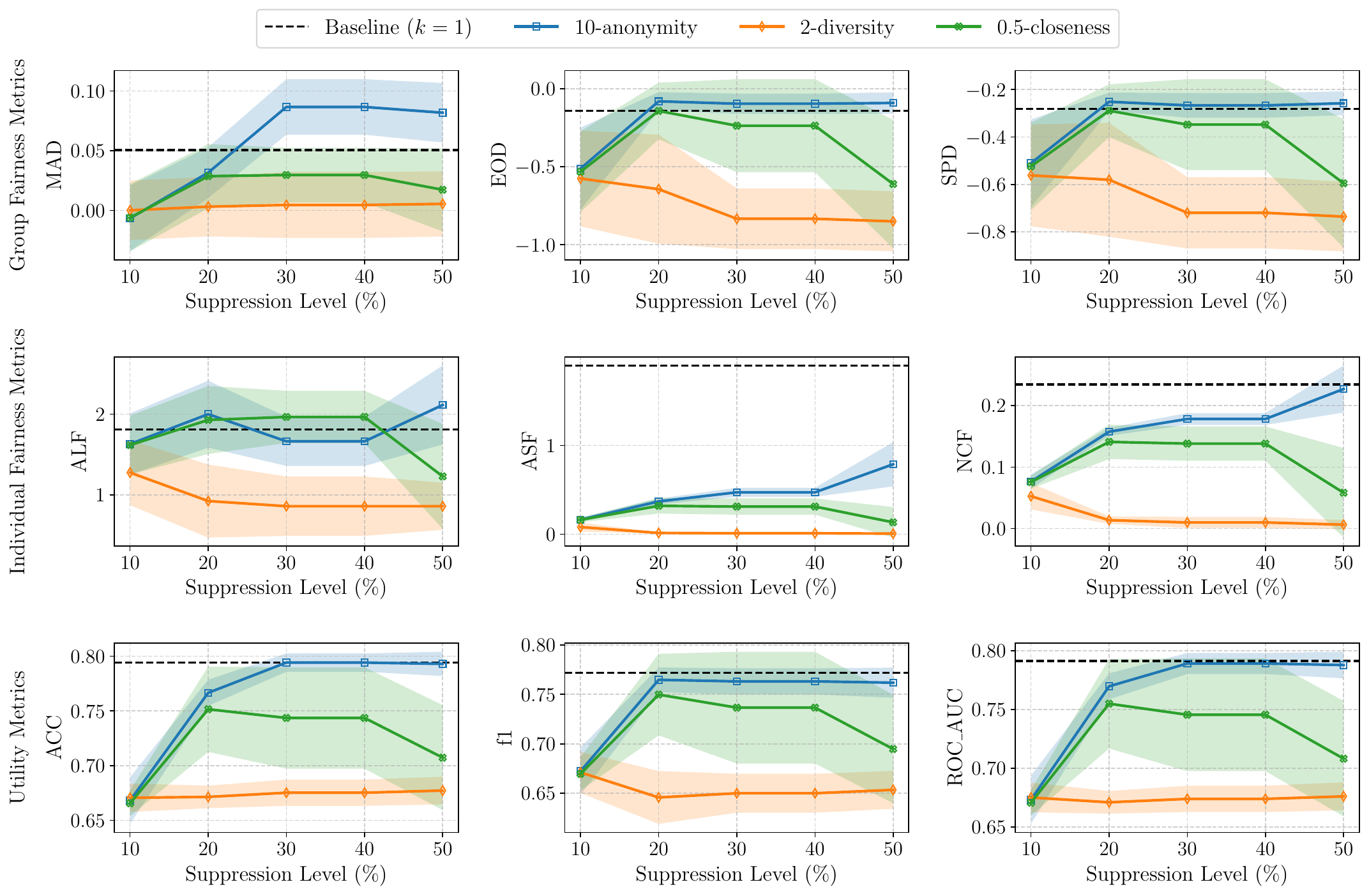}
    
    \caption{Impact of suppression level in anonymization (\kanon, \ldiv, \tclos) on group fairness (MAD, EOD, SPD), individual fairness (ALF, ASF, NCF), and utility (Accuracy, F1-score, ROC AUC) metrics in ML.
    Results with the \compas{} dataset with \race{} as the protected attribute for fairness evaluation.}
    \label{fig:suppression_results_compas_race}
\end{figure}

\begin{figure}[!htb]
    \centering
    
    \includegraphics[width=0.84\linewidth]{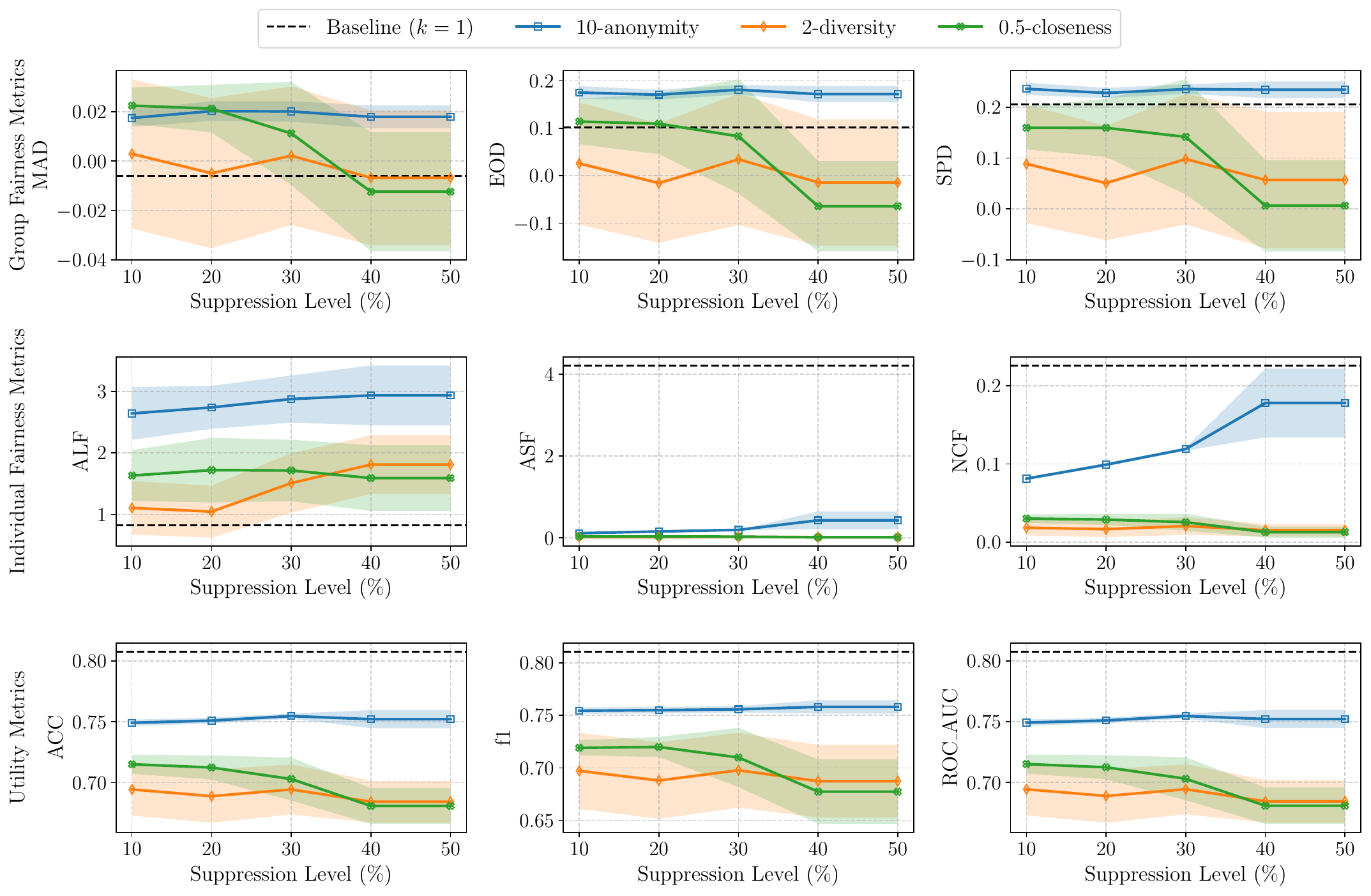}
    
    \caption{Impact of suppression level in anonymization (\kanon, \ldiv, \tclos) on group fairness (MAD, EOD, SPD), individual fairness (ALF, ASF, NCF), and utility (Accuracy, F1-score, ROC AUC) in ML.
    Results with the \acsincome{} dataset with \gender{} as the protected attribute for fairness evaluation.}
    \label{fig:suppression_results_acsincome_sex}
\end{figure}

\begin{figure}[!htb]
    \centering
    
    \includegraphics[width=0.84\linewidth]{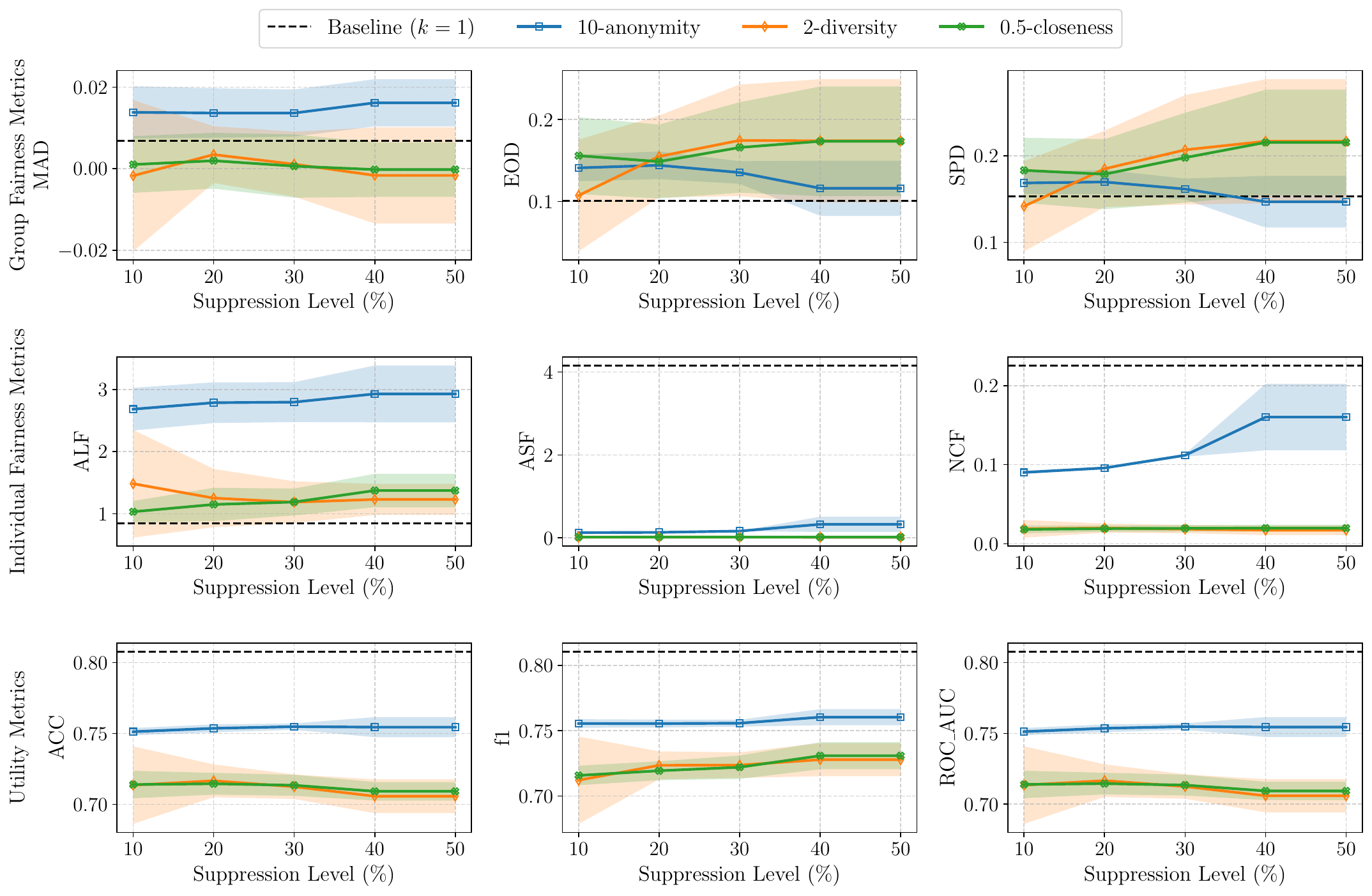}
    
    \caption{Impact of suppression level in anonymization (\kanon, \ldiv, \tclos) on group fairness (MAD, EOD, SPD), individual fairness (ALF, ASF, NCF), and utility (Accuracy, F1-score, ROC AUC) metrics in ML.
    Results with the \acsincome{} dataset with \race{} as the protected attribute for fairness evaluation.}
    \label{fig:suppression_results_acsincome_race}
\end{figure}

Figures~\ref{fig:suppression_results_adult_race}, \ref{fig:suppression_results_compas_sex}, \ref{fig:suppression_results_compas_race}, \ref{fig:suppression_results_acsincome_sex}, and \ref{fig:suppression_results_acsincome_race} demonstrate the impact of suppression levels on ML fairness for the \adult{} dataset with \race{} as the protected attribute and for both \compas{} and \acsincome{} datasets, considering \gender{} and \race{} as protected attributes, respectively. 
These results extend the experiments presented in Section~\ref{sub:results_suppression}.

It can be observed that the impact of suppression levels on fairness and utility metrics exhibits complex patterns across datasets and protected attributes. For group fairness, metrics such as SPD, EOD, and MAD show mixed trends—moderate suppression levels sometimes lead to slight improvements by removing outliers, while in other cases, especially in the \acsincome{} dataset, fairness metrics remain stable or worsen due to the introduction of demographic disparities. In contrast, individual fairness metrics (ALF, ASF, and NCF) consistently degrade as suppression increases, as the removal of records disrupts local consistency in predictions. Regarding utility, suppression generally has a neutral or slightly positive effect by eliminating noisy data, but at high levels, it can reduce model performance, particularly for \ldiv{} and \tclos.

\subsection{ML Fairness Across Target Distribution Variations}

\begin{figure}[!htb]
    \centering
    
    \includegraphics[width=0.84\linewidth]{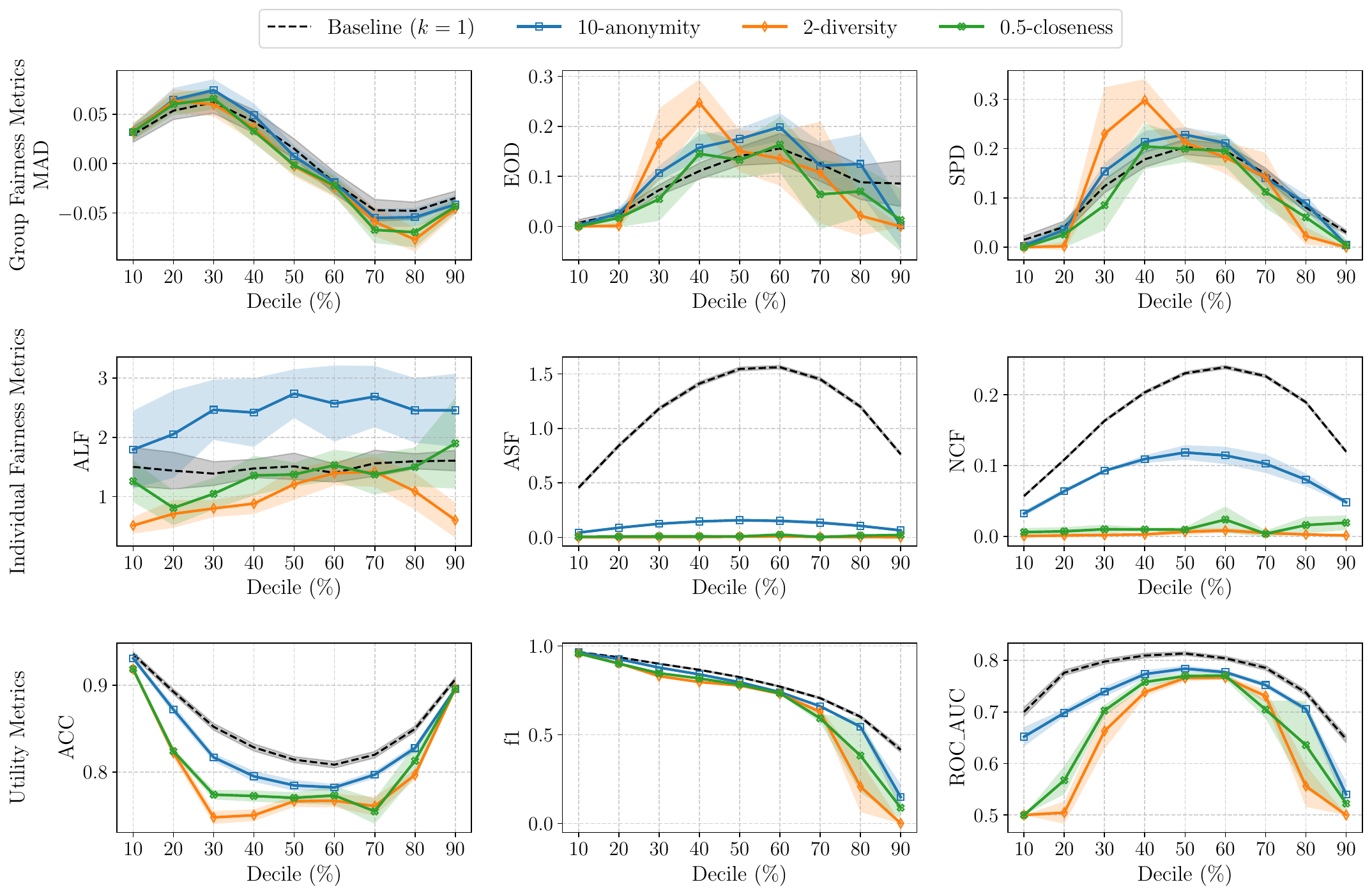}
        
    \caption{Effect of target distribution changes (\ie, thresholded at deciles ranging from 10\% to 90\%) on anonymity techniques (\kanon, \ldiv, \tclos) regarding group fairness (MAD, EOD, SPD), individual fairness (ALF, ASF, NCF), and utility (Accuracy, F1-score, ROC AUC) metrics in ML. Results are presented for the \adult{} dataset, with \race{} serving as the protected attribute for fairness evaluation.}
    \label{fig:target_distribution_results_adult_race}
\end{figure}

\begin{figure}[!htb]
    \centering
    
    \includegraphics[width=0.84\linewidth]{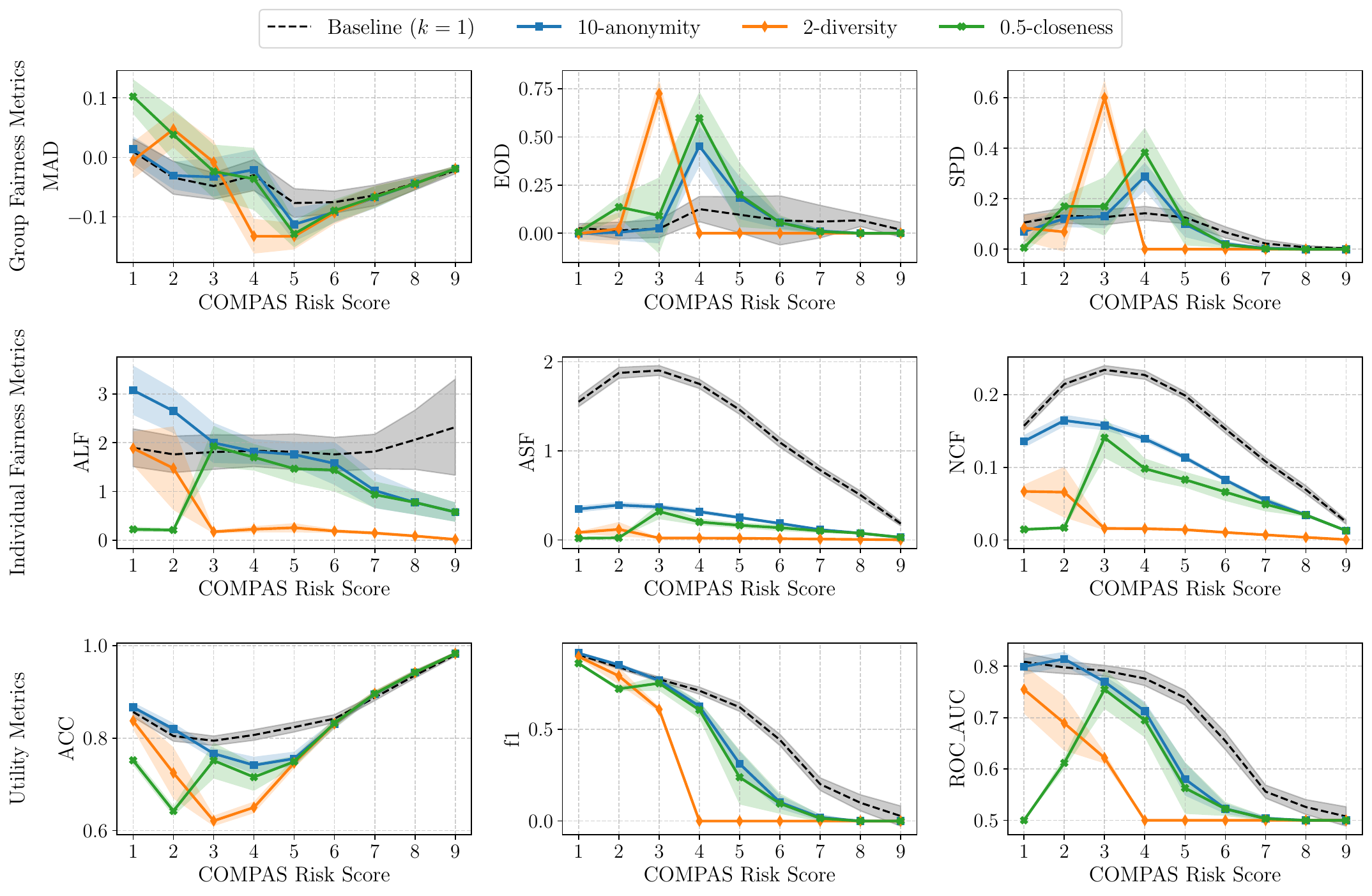}
        
    \caption{Effect of target distribution changes (\ie, thresholded at COMPAS Risk Score ranging from 1 to 9) on anonymity techniques (\kanon, \ldiv, \tclos) regarding group fairness (MAD, EOD, SPD), individual fairness (ALF, ASF, NCF), and utility (Accuracy, F1-score, ROC AUC) metrics in ML. Results are presented for the \compas{} dataset, with \gender{} serving as the protected attribute for fairness evaluation.}
    \label{fig:target_distribution_results_compas_sex}
\end{figure}

\begin{figure}[!htb]
    \centering
    
    \includegraphics[width=0.84\linewidth]{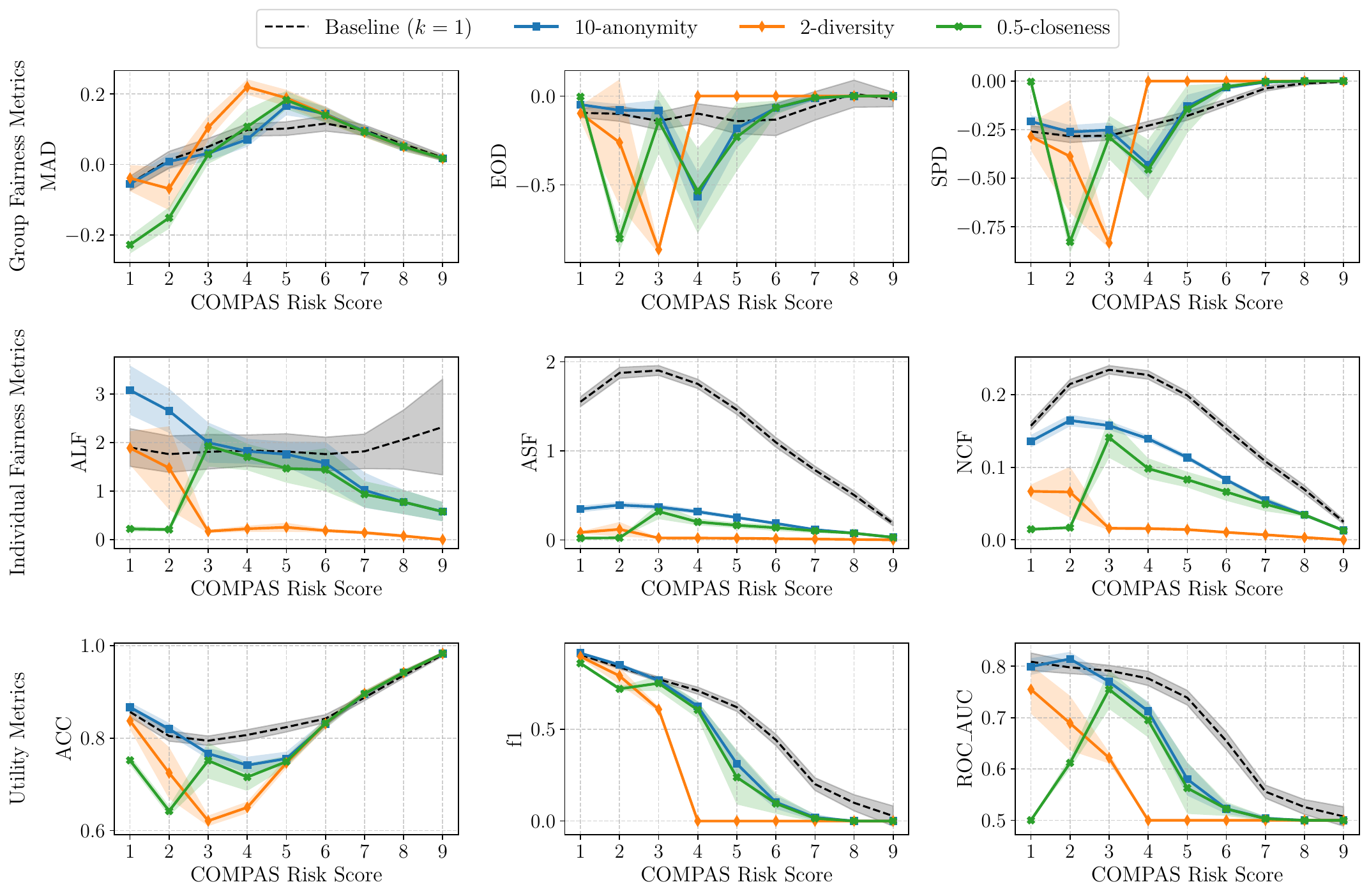}    
        
    \caption{Effect of target distribution changes (\ie, thresholded at COMPAS Risk Score ranging from 1 to 9) on anonymity techniques (\kanon, \ldiv, \tclos) regarding group fairness (MAD, EOD, SPD), individual fairness (ALF, ASF, NCF), and utility (Accuracy, F1-score, ROC AUC) metrics in ML. Results are presented for the \compas{} dataset, with \race{} serving as the protected attribute for fairness evaluation.}
    \label{fig:target_distribution_results_compas_race}
\end{figure}

\begin{figure}[!htb]
    \centering
    
    \includegraphics[width=0.84\linewidth]{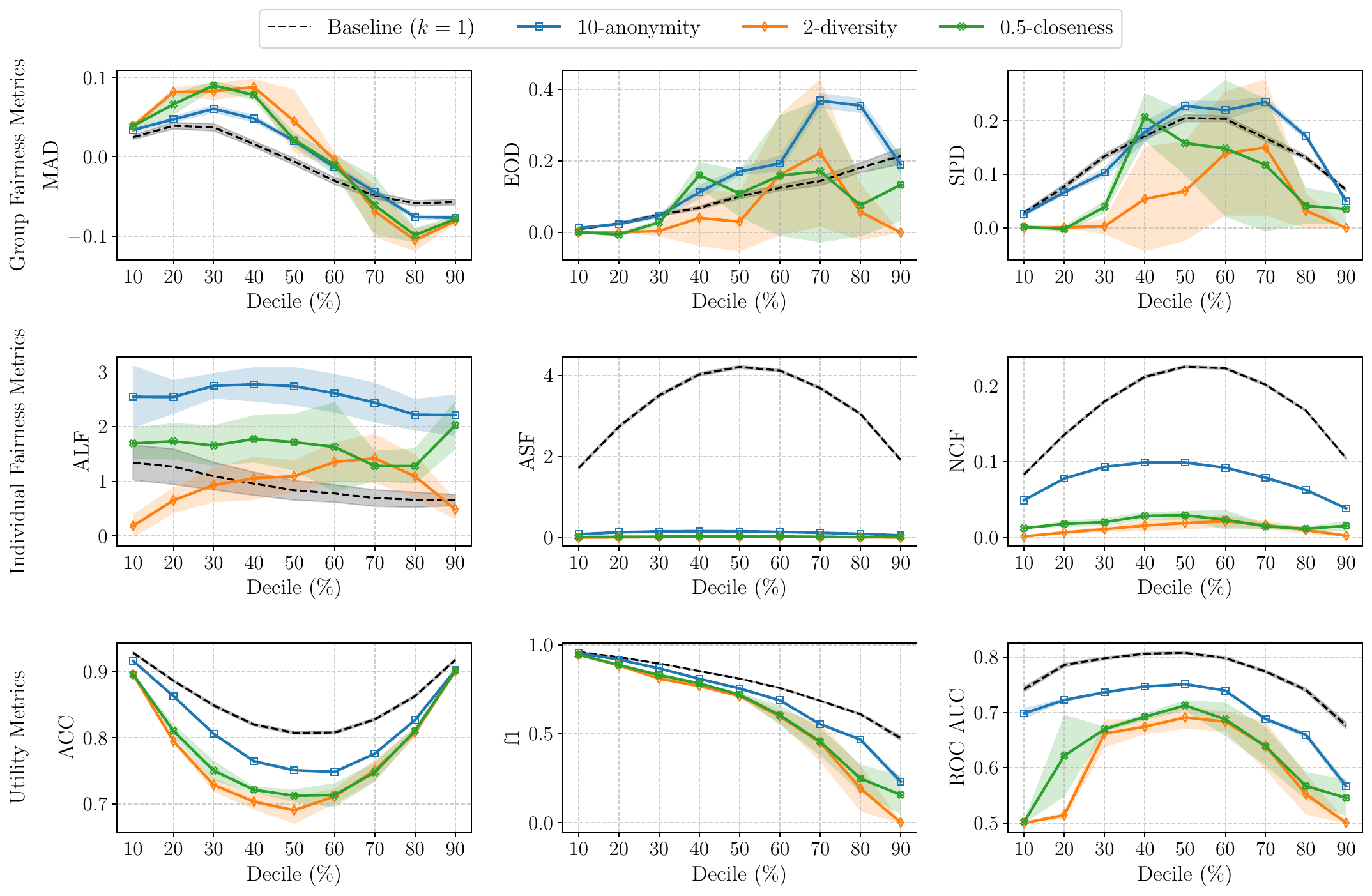}
        
    \caption{Effect of target distribution changes (\ie, thresholded at deciles ranging from 10\% to 90\%) on anonymity techniques (\kanon, \ldiv, \tclos) regarding group fairness (MAD, EOD, SPD), individual fairness (ALF, ASF, NCF), and utility (Accuracy, F1-score, ROC AUC) metrics in ML. Results are presented for the \acsincome{} dataset, with \gender{} serving as the protected attribute for fairness evaluation.}
    \label{fig:target_distribution_results_ACSIncome_SEX}
\end{figure}

\begin{figure}[!htb]
    \centering
    
    \includegraphics[width=0.84\linewidth]{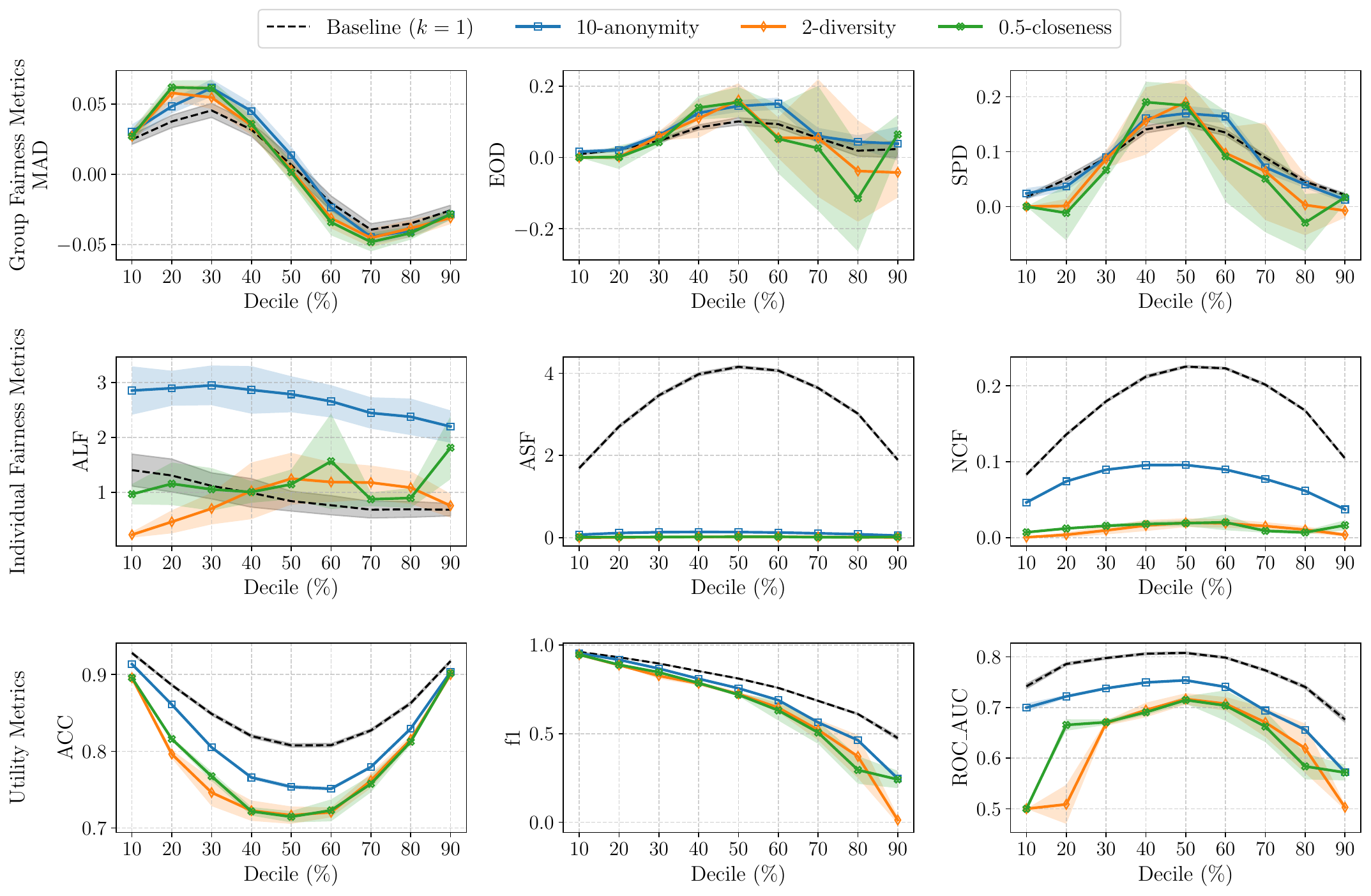}    
        
    \caption{Effect of target distribution changes (\ie, thresholded at deciles ranging from 10\% to 90\%) on anonymity techniques (\kanon, \ldiv, \tclos) regarding group fairness (MAD, EOD, SPD), individual fairness (ALF, ASF, NCF), and utility (Accuracy, F1-score, ROC AUC) metrics in ML. Results are presented for the \acsincome{} dataset, with \race{} serving as the protected attribute for fairness evaluation.}
    \label{fig:target_distribution_results_ACSIncome_race}
\end{figure}

Figures~\ref{fig:target_distribution_results_adult_race}, \ref{fig:target_distribution_results_compas_sex}, 
\ref{fig:target_distribution_results_compas_race}, 
\ref{fig:target_distribution_results_ACSIncome_SEX}, and \ref{fig:target_distribution_results_ACSIncome_race} show the ML fairness across target distribution for the \adult{} dataset with \race{} as the protected attribute and for both \compas{} and \acsincome{} datasets, considering \gender{} and \race{} as protected attributes, respectively. 
These results extend the experiments presented in Section~\ref{sub:results_target_distribution}. 

These results reveal that anonymization techniques generally preserve the overall distributional patterns of fairness and utility metrics compared to the baseline. However, they significantly affect metric magnitudes: group fairness metrics (MAD, SPD, and EOD) deteriorate under anonymization, amplifying disparities between demographic groups, whereas individual fairness metrics (ASF and NCF) tend to improve, indicating greater consistency in model predictions. Meanwhile, utility metrics (ACC, F1-score, and ROC AUC) consistently decline, highlighting the trade-off between privacy and predictive performance. 

\subsection{Impact of Data Size on Fairness in ML}

\begin{figure}[!htb]
    \centering
    
    \includegraphics[width=0.84\linewidth]{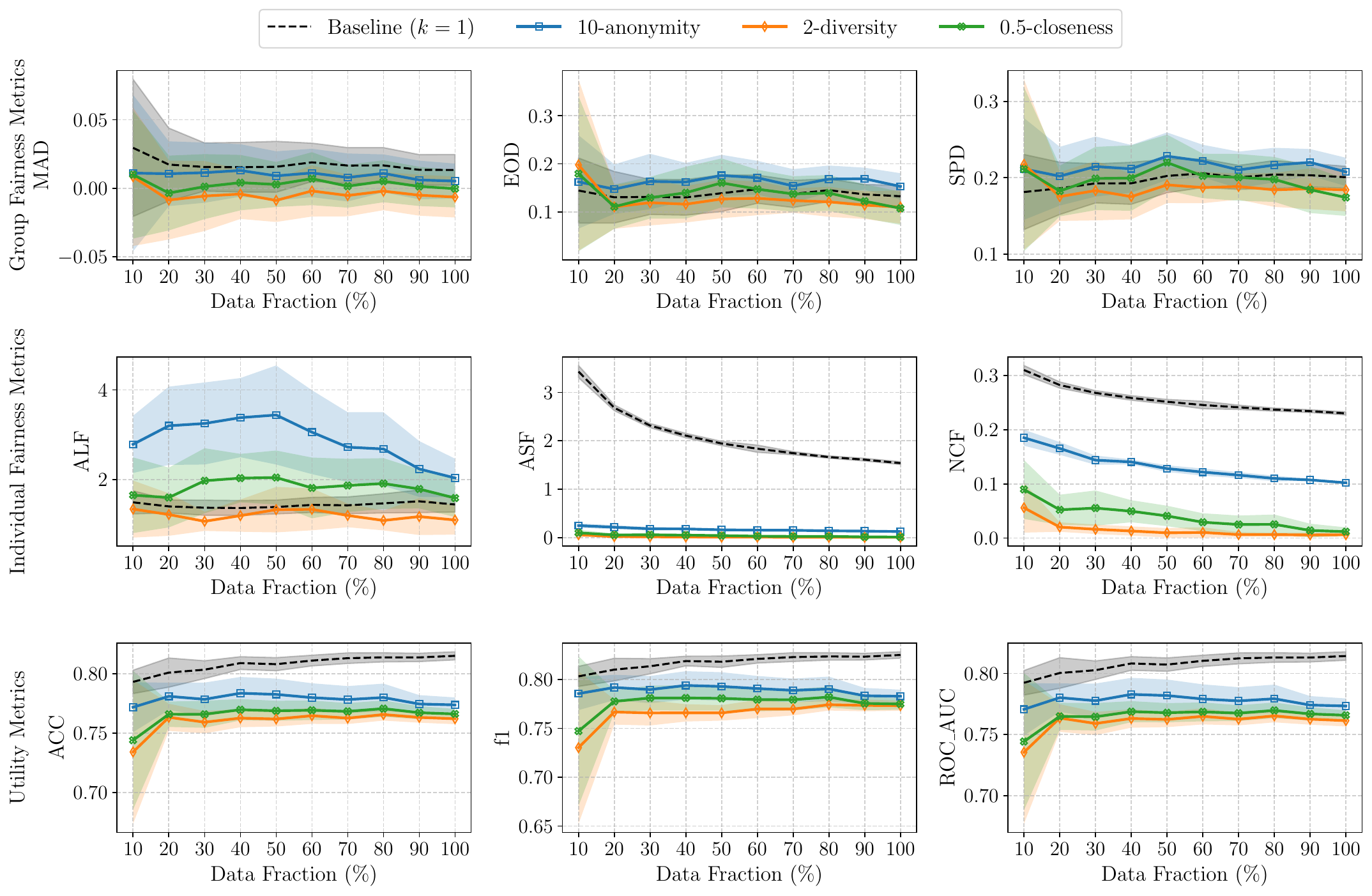}
    
    \caption{Effect of varying data fraction on the performance of anonymity techniques ($10$-anonymity, $2$-diversity, $0.5$-closeness) in terms of group fairness (MAD, EOD, SPD), individual fairness (ALF, ASF, NCF), and utility (Accuracy, F1-score, ROC AUC) metrics in ML. This analysis is performed using the \adult{} dataset, considering \race{} as the protected attribute for fairness evaluation.}
    \label{fig:data_fraction_results_adult_race}
\end{figure}

\begin{figure}[!htb]
    \centering
    
    \includegraphics[width=0.84\linewidth]{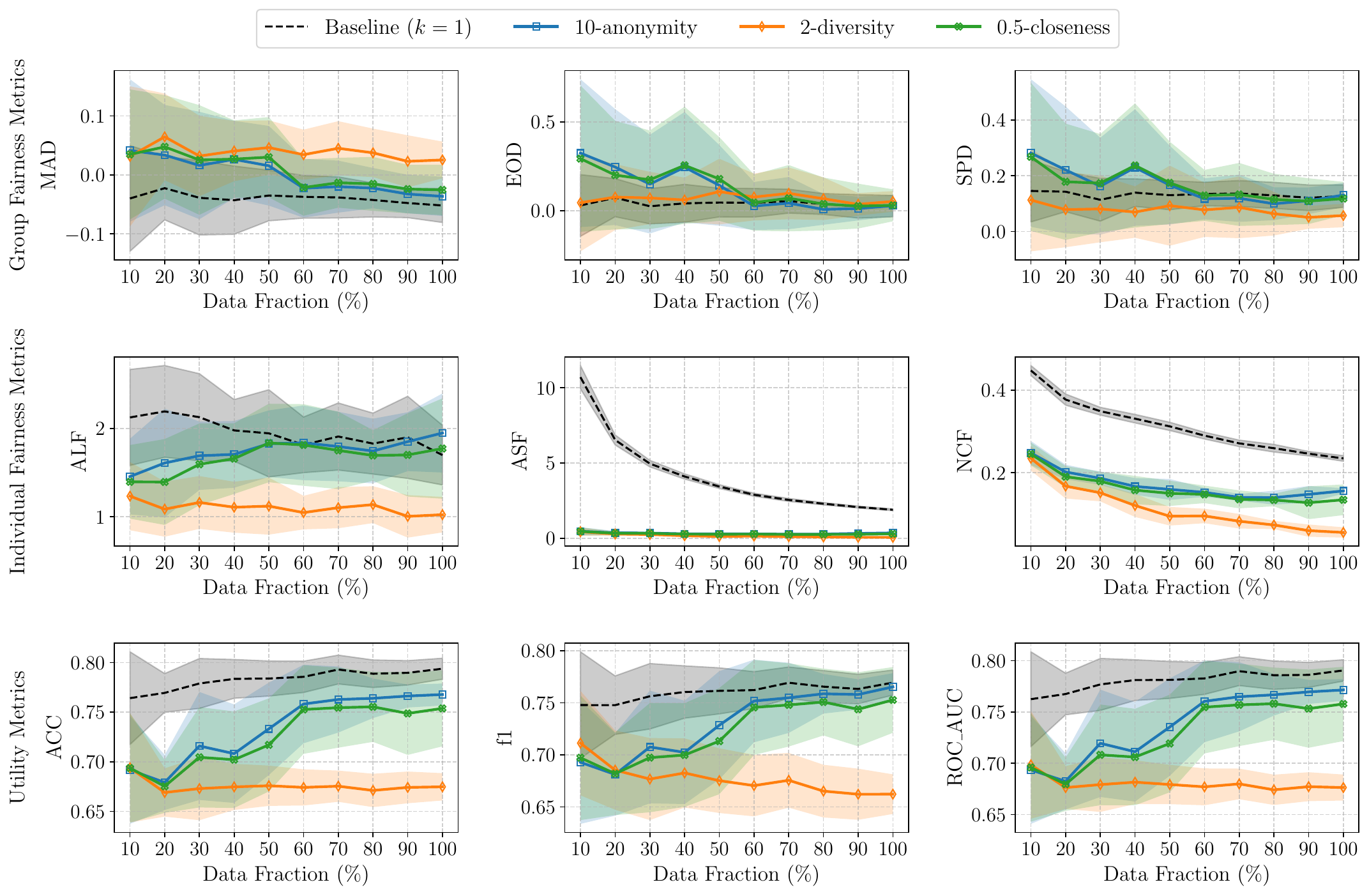}
    
    \caption{Effect of varying data fraction on the performance of anonymity techniques ($10$-anonymity, $2$-diversity, $0.5$-closeness) in terms of group fairness (MAD, EOD, SPD), individual fairness (ALF, ASF, NCF), and utility (Accuracy, F1-score, ROC AUC) metrics in ML. This analysis is performed using the \compas{} dataset, considering \gender{} as the protected attribute for fairness evaluation.}
    \label{fig:data_fraction_results_compas_sex}
\end{figure}

\begin{figure}[!htb]
    \centering
    
    \includegraphics[width=0.84\linewidth]{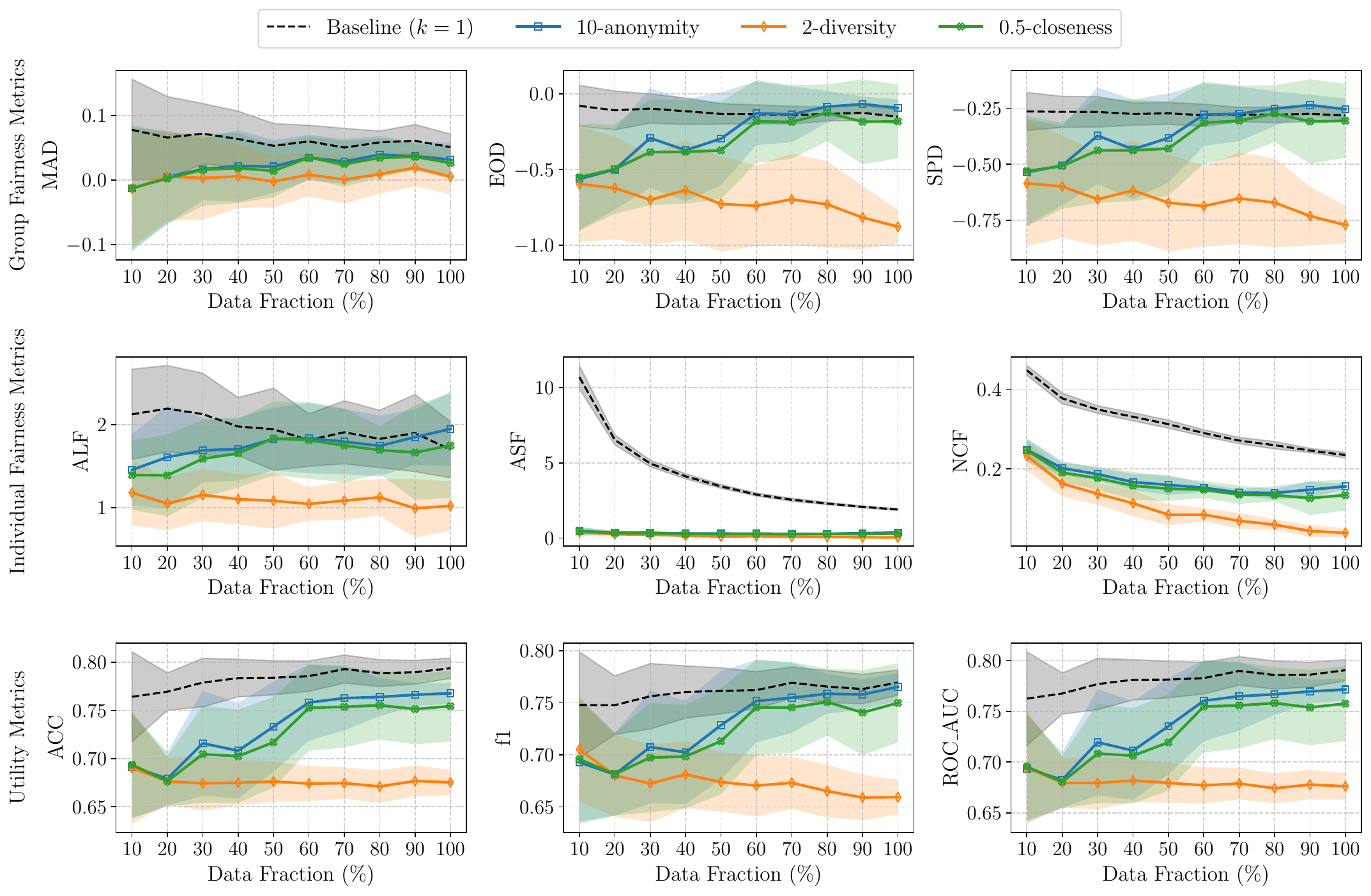}
    
    \caption{Effect of varying data fraction on the performance of anonymity techniques ($10$-anonymity, $2$-diversity, $0.5$-closeness) in terms of group fairness (MAD, EOD, SPD), individual fairness (ALF, ASF, NCF), and utility (Accuracy, F1-score, ROC AUC) metrics in ML. This analysis is performed using the \compas{} dataset, considering \race{} as the protected attribute for fairness evaluation.}
    \label{fig:data_fraction_results_compas_race}
\end{figure}

\begin{figure}[!htb]
    \centering
    
    \includegraphics[width=0.84\linewidth]{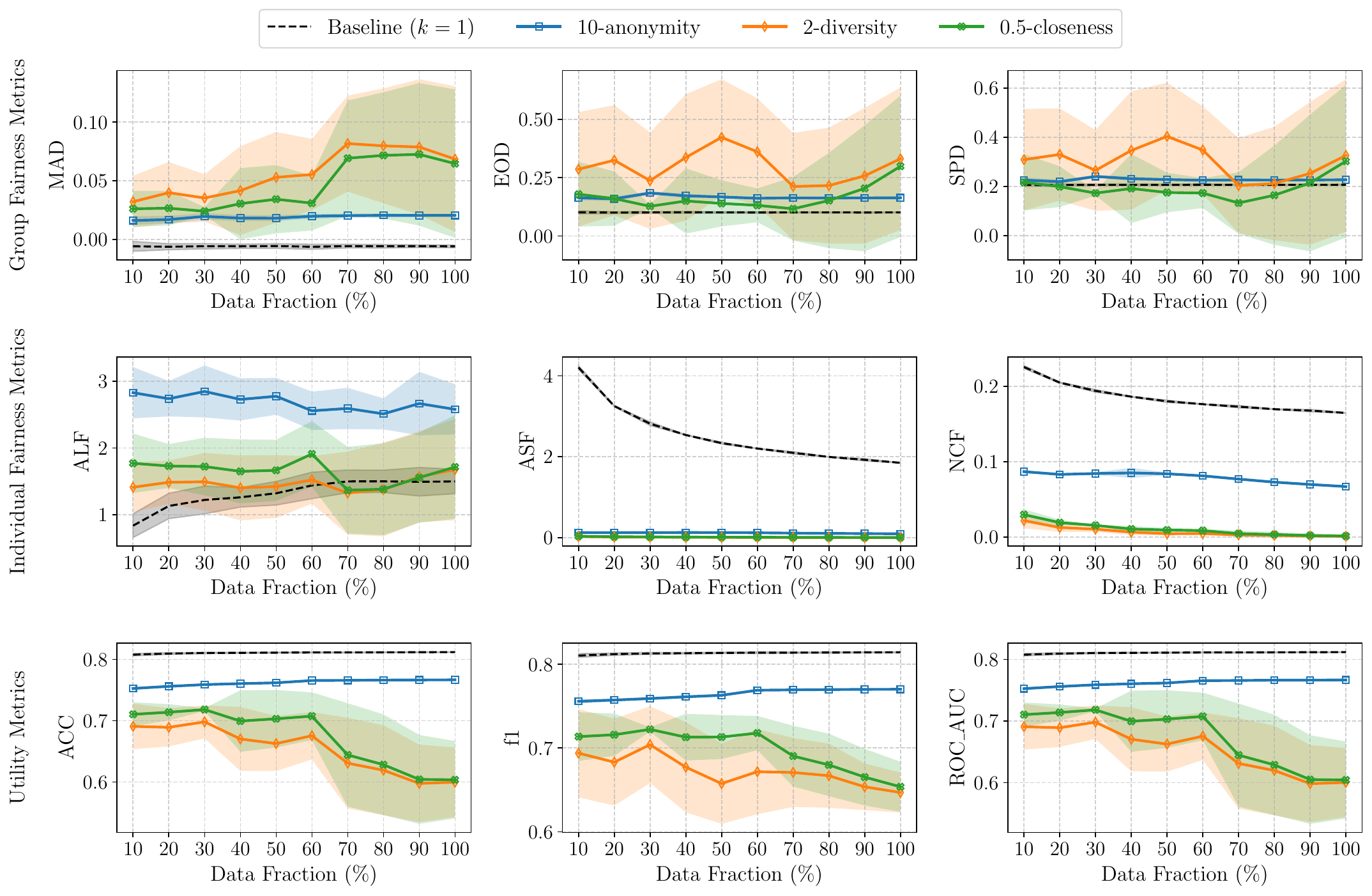}
    
    \caption{Effect of varying data fraction on the performance of anonymity techniques ($10$-anonymity, $2$-diversity, $0.5$-closeness) in terms of group fairness (MAD, EOD, SPD), individual fairness (ALF, ASF, NCF), and utility (Accuracy, F1-score, ROC AUC) metrics in ML. This analysis is performed using the \acsincome{} dataset, considering \gender{} as the protected attribute for fairness evaluation.}
    \label{fig:data_fraction_results_ACSIncome_gender}
\end{figure}

\begin{figure}[!htb]
    \centering
    
    \includegraphics[width=0.84\linewidth]{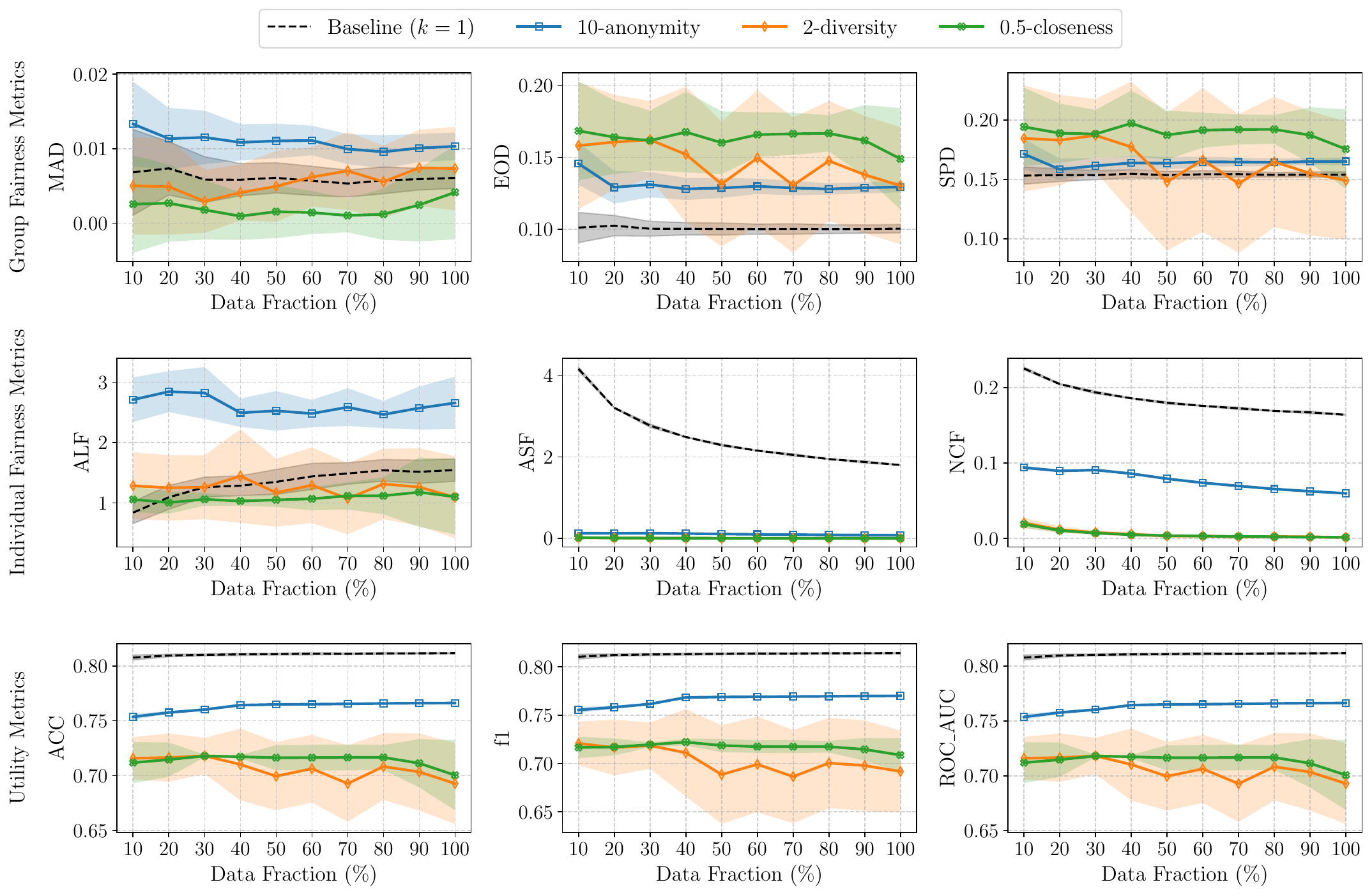}
    
    \caption{Effect of varying data fraction on the performance of anonymity techniques ($10$-anonymity, $2$-diversity, $0.5$-closeness) in terms of group fairness (MAD, EOD, SPD), individual fairness (ALF, ASF, NCF), and utility (Accuracy, F1-score, ROC AUC) metrics in ML. This analysis is performed using the \acsincome{} dataset, considering \race{} as the protected attribute for fairness evaluation.}
    \label{fig:data_fraction_results_ACSIncome_race}
\end{figure}

Figures~\ref{fig:data_fraction_results_adult_race},  \ref{fig:data_fraction_results_compas_sex}, 
\ref{fig:data_fraction_results_compas_race}, \ref{fig:data_fraction_results_ACSIncome_gender}, and~\ref{fig:data_fraction_results_ACSIncome_race} show the effect of dataset sizes on ML fariness for the \adult{} dataset with \race{} as the protected attribute and for both \compas{} and \acsincome{} datasets, considering \gender{} and \race{} as protected attributes, respectively. 
These results extend the experiments presented in Section~\ref{sub:results_data_size}. 

These experiments demonstrate that fairness and performance metrics under anonymization remain relatively stable across varying data fractions, ranging from 10\% to 100\% of the dataset. These results indicate that the trade-offs between privacy, fairness, and utility are predominantly influenced by the choice of anonymization techniques and their parameter configurations, rather than by the dataset size. While random sampling introduces slight variations, it does not fundamentally alter the observed trends, reinforcing the robustness of the identified patterns. 

\subsection{Comparison of ML Classifiers in Fairness Under Anonymization}

\begin{figure}[!htb]
    \centering
    
    \includegraphics[width=0.84\linewidth]{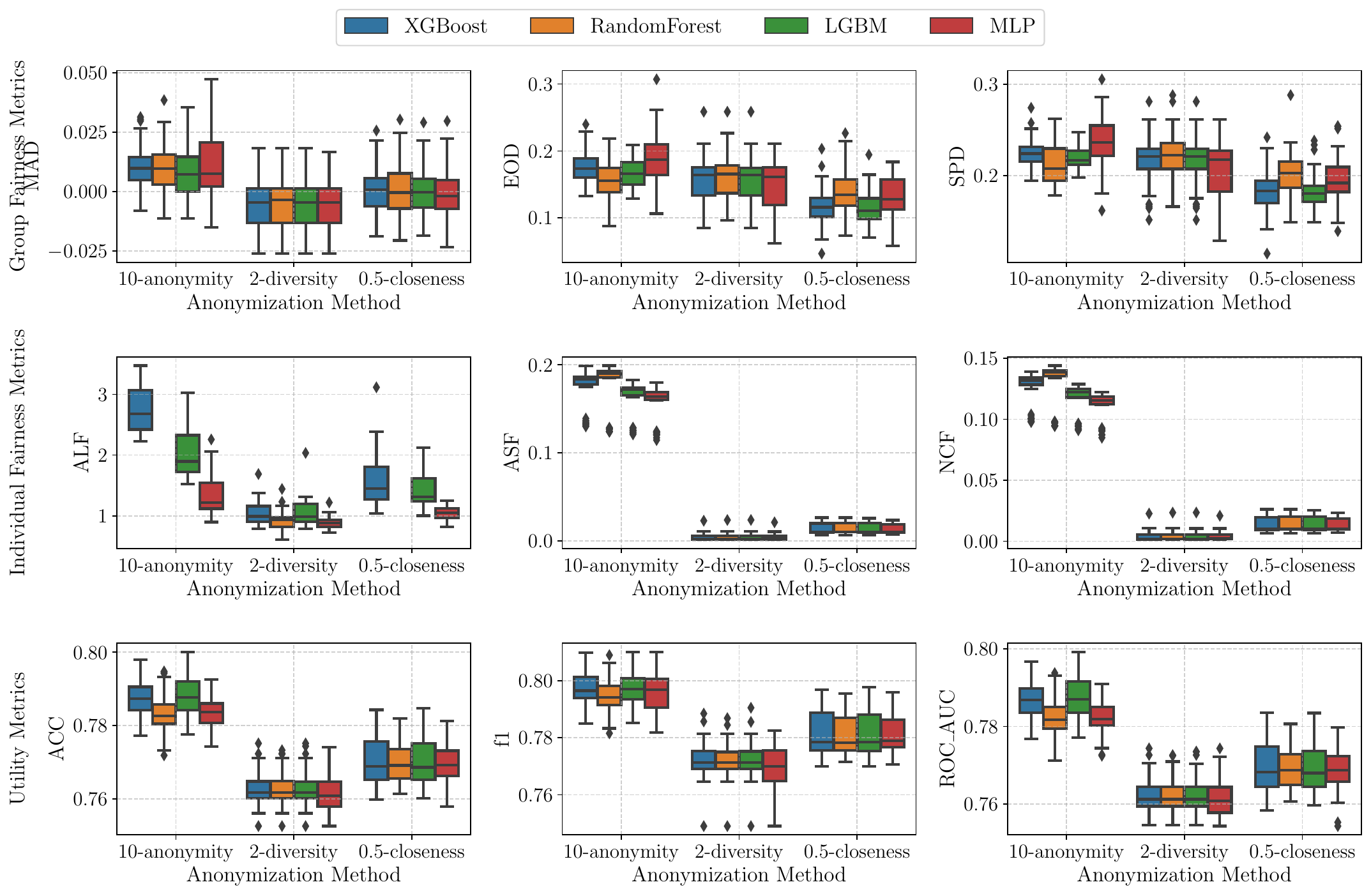}
    
    \caption{Comparison of the impact of different state-of-the-art ML classifiers on anonymized dataset (\kanon, \ldiv, \tclos) and relation to group fairness (MAD, EOD, SPD), individual fairness (ALF, ASF, NCF), and utility (Accuracy, F1-score, ROC AUC) metrics in ML. Results are based on the \adult{} dataset, with \race{} as the protected attribute for fairness evaluation.}
    \label{fig:classifier_results_adult_race}
\end{figure}

\begin{figure}[!htb]
    \centering
    
    \includegraphics[width=0.84\linewidth]{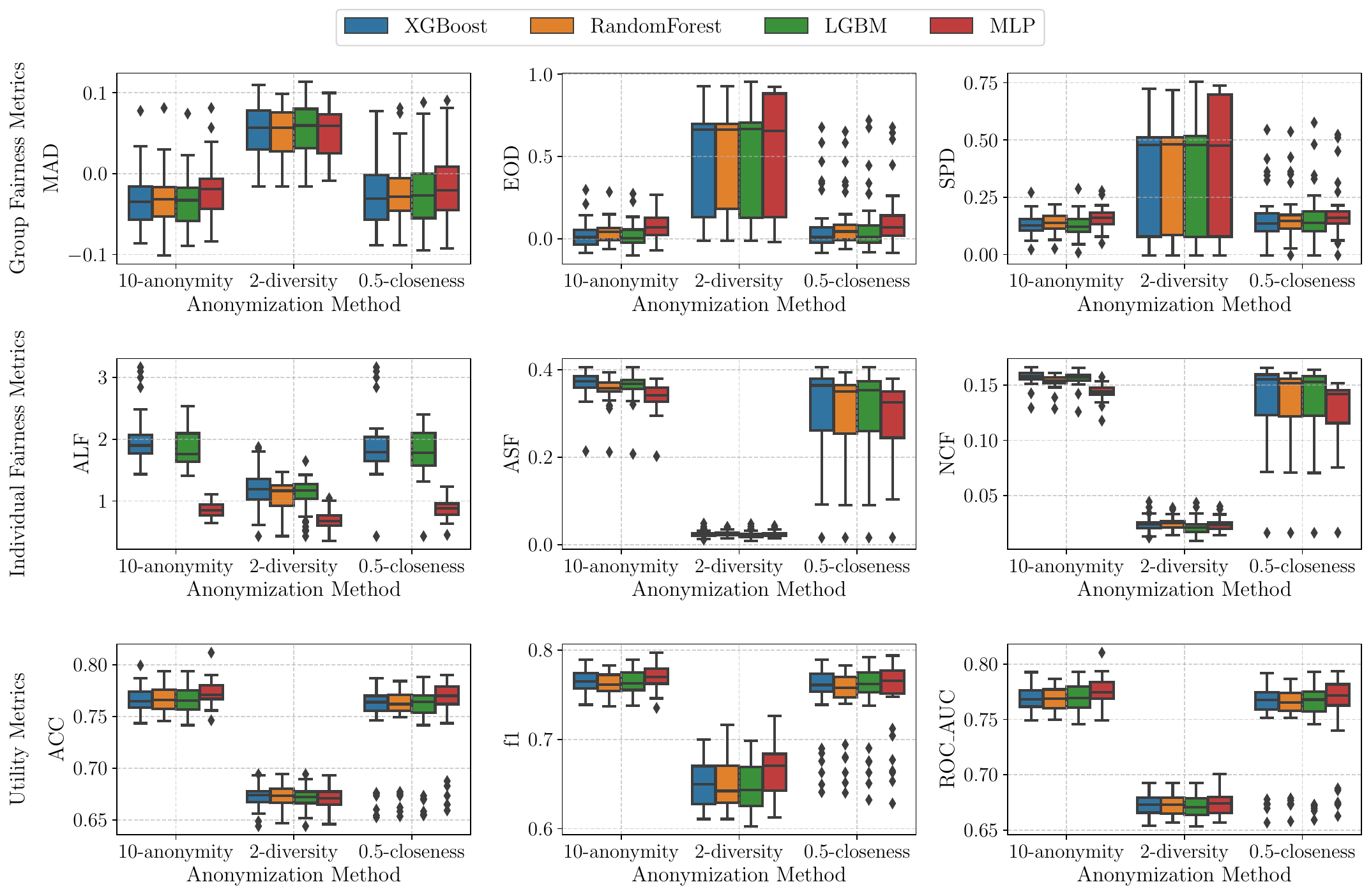}
    
    \caption{Comparison of the impact of different state-of-the-art ML classifiers on anonymized dataset (\kanon, \ldiv, \tclos) and relation to group fairness (MAD, EOD, SPD), individual fairness (ALF, ASF, NCF), and utility (Accuracy, F1-score, ROC AUC) metrics in ML. Results are based on the \compas{} dataset, with \gender{} as the protected attribute for fairness evaluation.}
    \label{fig:classifier_results_compas_sex}
\end{figure}

\begin{figure}[!htb]
    \centering
    
    \includegraphics[width=0.84\linewidth]{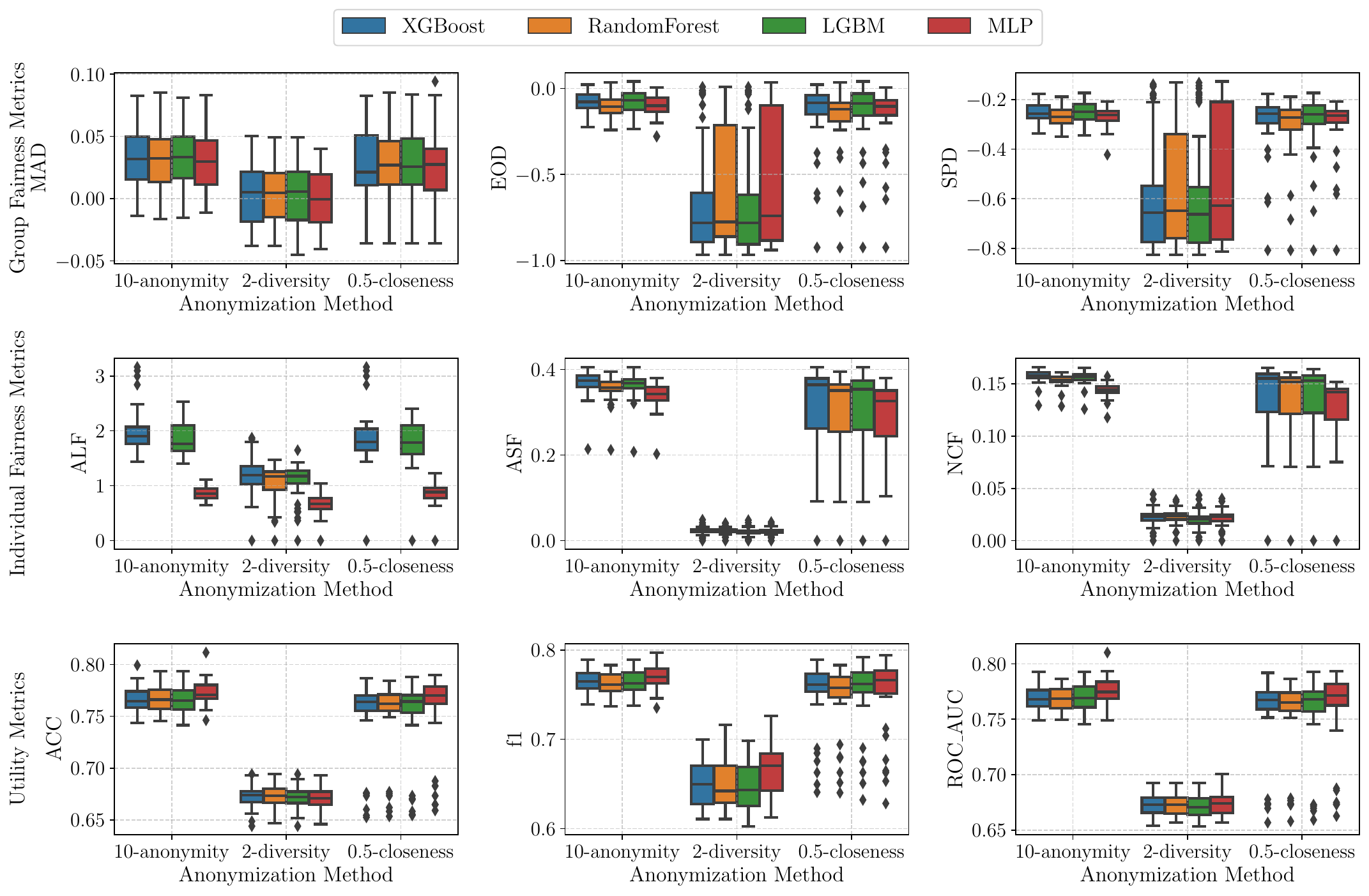}
    
    \caption{Comparison of the impact of different state-of-the-art ML classifiers on anonymized dataset (\kanon, \ldiv, \tclos) and relation to group fairness (MAD, EOD, SPD), individual fairness (ALF, ASF, NCF), and utility (Accuracy, F1-score, ROC AUC) metrics in ML. Results are based on the \compas{} dataset, with \race{} as the protected attribute for fairness evaluation.}
    \label{fig:classifier_results_compas_race}
\end{figure}

\begin{figure}[!htb]
    \centering
    
    \includegraphics[width=0.84\linewidth]{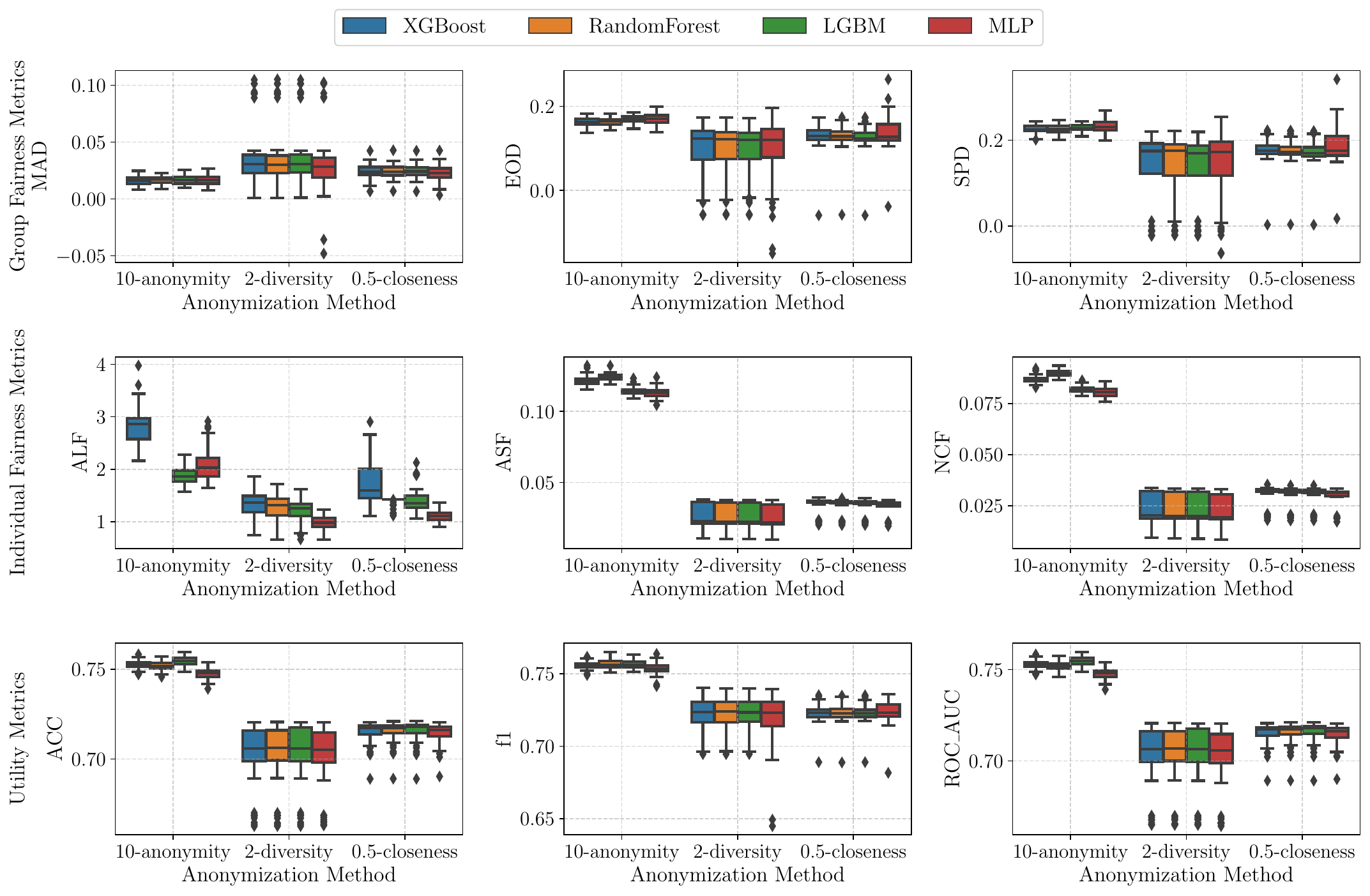}
    
    \caption{Comparison of the impact of different state-of-the-art ML classifiers on anonymized dataset (\kanon, \ldiv, \tclos) and relation to group fairness (MAD, EOD, SPD), individual fairness (ALF, ASF, NCF), and utility (Accuracy, F1-score, ROC AUC) metrics in ML. Results are based on the \acsincome{} dataset, with \gender{} as the protected attribute for fairness evaluation.}
    \label{fig:classifier_results_ASCIncome_SEX}
\end{figure}

\begin{figure}[!htb]
    \centering
    
    \includegraphics[width=0.84\linewidth]{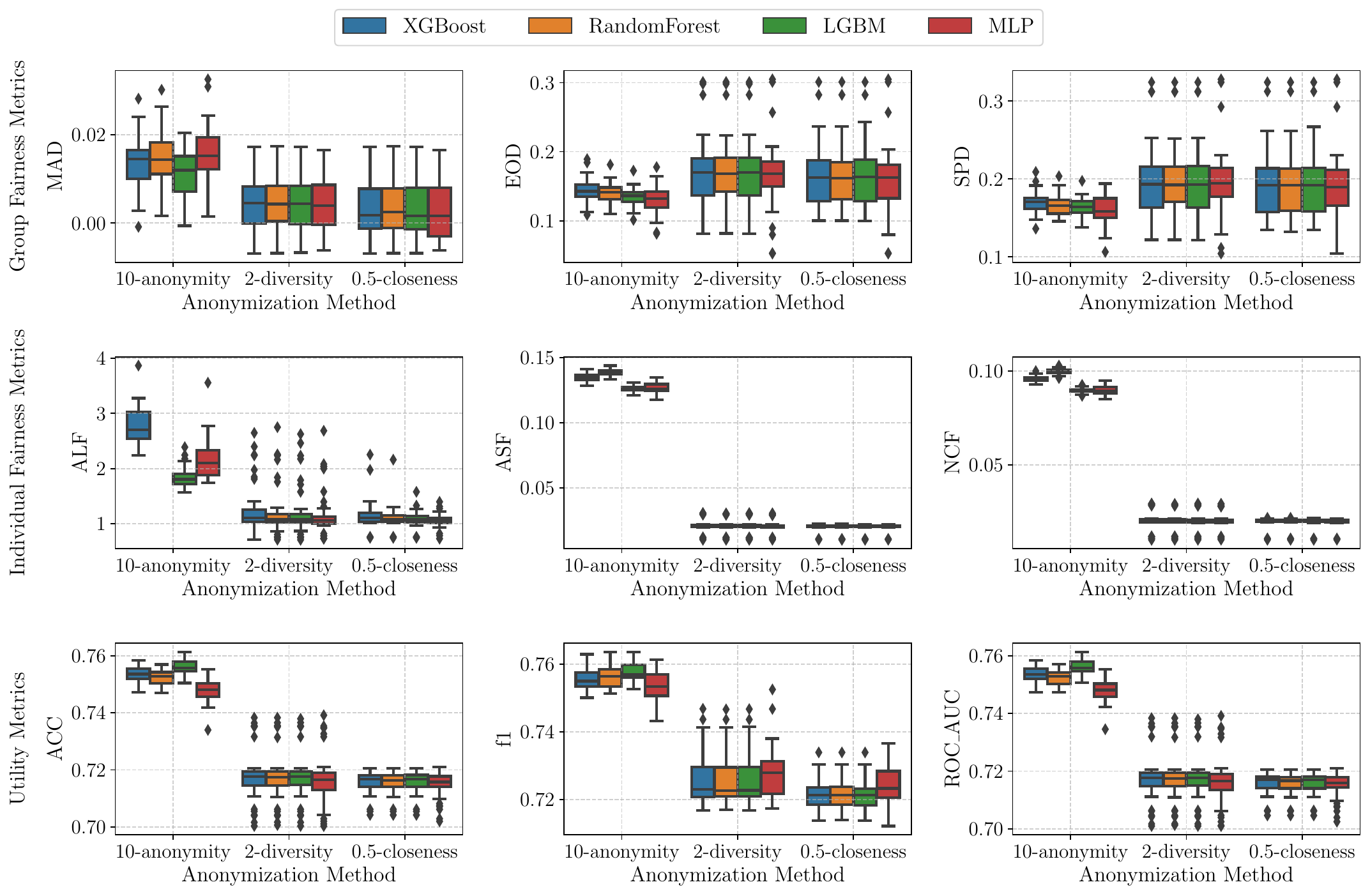}
    
    \caption{Comparison of the impact of different state-of-the-art ML classifiers on anonymized dataset (\kanon, \ldiv, \tclos) and relation to group fairness (MAD, EOD, SPD), individual fairness (ALF, ASF, NCF), and utility (Accuracy, F1-score, ROC AUC) metrics in ML. Results are based on the \acsincome{} dataset, with \race{} as the protected attribute for fairness evaluation.}
    \label{fig:classifier_results_ACSIncome_race}
\end{figure}

Figures~\ref{fig:classifier_results_adult_race},~\ref{fig:classifier_results_compas_sex},~\ref{fig:classifier_results_compas_race},~\ref{fig:classifier_results_ASCIncome_SEX}, and~\ref{fig:classifier_results_ACSIncome_race} show the ML fairness using different ML models on anonymizaed datasets for the \adult{} dataset with \race{} as the protected attribute and for the \acsincome{} dataset, considering \texttt{SEX} and \texttt{RAC1P} as protected attributes, respectively. 
These results extend the experiments presented in Section~\ref{sub:results_ml_classifier}. 

The results confirm that the trends observed with XGBoost remain consistent across different classifiers. While minor variations exist, such as XGBoost exhibiting slightly better utility and fairness stability, the overall patterns persist. Specifically, anonymization continues to negatively affect group fairness, improve individual fairness, and introduce trade-offs in utility across classifiers. These findings indicate that the insights gained from XGBoost-based experiments are broadly applicable to other models, reinforcing the generalizability of our study's conclusions.

\end{document}